\pdfoutput=1

\documentclass[11pt]{article}

\usepackage[]{acl}

\usepackage{times}
\usepackage{latexsym}

\usepackage[T1]{fontenc}

\usepackage[utf8]{inputenc}

\usepackage{microtype}
\usepackage{microtype}
\usepackage{multirow}
\usepackage{stfloats}
\usepackage{xcolor,colortbl}
\usepackage{times}
\usepackage{latexsym}
\usepackage{graphicx}
\usepackage{graphics}
\usepackage{multirow}
\usepackage{amsmath}
\usepackage{tabularray}
\usepackage{longtable}
\usepackage{supertabular,booktabs}
\usepackage{multicol}
\usepackage{amsfonts}
\usepackage{amssymb}
\usepackage{tablefootnote}
\usepackage{color}
\usepackage{bbm}
\usepackage{graphicx}
\usepackage{xspace}
\PassOptionsToPackage{hyphens}{url}
\usepackage{hyperref}
\usepackage{url}
\usepackage[export]{adjustbox}

\usepackage{subcaption}
\usepackage{fdsymbol}
\usepackage{cleveref}
\usepackage{caption}
\usepackage{booktabs}
\usepackage{makecell}
\usepackage{multirow}
\usepackage{adjustbox}
\usepackage{array}

\usepackage{enumitem}
\newcolumntype{P}[1]{>{\centering\arraybackslash}p{#1}}

\newcommand{\fc}{{factually consistent}}
\newcommand{\fic}{{factually inconsistent}}
\newcommand{\benchmark}{{\fontfamily{lmtt}\selectfont FIB}\xspace}

\title{Evaluating the Factual Consistency of Large Language Models \\Through News Summarization}

\author{
 Derek Tam \quad Anisha Mascarenhas \quad Shiyue Zhang \AND Sarah Kwan \quad Mohit Bansal \quad Colin Raffel \vspace{0.5em}\\
 University of North Carolina at Chapel Hill \\
\texttt{ \{dtredsox,amascare,shiyue,mbansal,craffel\}@cs.unc.edu}
}

\begin{document}
\maketitle
\begin{abstract}
While large language models (LLMs) have proven to be effective on a large variety of tasks, they are also known to hallucinate information. 
To measure whether an LLM prefers \fc{} continuations of its input, we propose a new benchmark called \benchmark(\textbf{F}actual \textbf{I}nconsistency \textbf{B}enchmark) that focuses on the task of summarization.
Specifically, our benchmark involves comparing the scores an LLM assigns to a \fc{} versus a \fic{} summary for an input news article.
For \fc{} summaries, we use human-written reference summaries that we manually verify as \fc{}.
To generate summaries that are \fic{}, we generate summaries from a suite of summarization models that we have manually annotated as \fic{}.
A model's factual consistency is then measured according to its accuracy, i.e.\ the proportion of documents where it assigns a higher score to the \fc{} summary.
To validate the usefulness of \benchmark, we evaluate 23 large language models ranging from 1B to 176B parameters from six different model families including BLOOM and OPT.
We find that existing LLMs generally assign a higher score to \fc{} summaries than to \fic{} summaries. 
However, if the \fic{} summaries occur verbatim in the document, then LLMs assign a higher score to these \fic{} summaries than \fc{} summaries.
We validate design choices in our benchmark including the scoring method and source of distractor summaries.
\footnote{We include our code in the supplementary}

\end{abstract}

\section{Introduction}
Factual inconsistency is a widespread problem in natural language generation tasks~\cite{maynez-etal-2020-faithfulness, weng-etal-2020-towards, devaraj-etal-2022-evaluating}. 
For text summarization in particular, it has been shown that models often hallucinate new information or generate content that contradicts the source document~\cite{cao2018faithful, maynez-etal-2020-faithfulness}. 
These works usually study supervised summarization models that are either trained from scratch or fine-tuned from a pre-trained language model \cite{wan-bansal-2022-factpegasus}.
Recently, however, NLP has experienced a paradigm shift towards using large language models (LLMs) rather than supervised models.
LLMs are generally pre-trained on a large corpus of unstructured text and then applied to a task through instructive prompts.
In light of this new paradigm, our goal is to evaluate the factual consistency of large language models using text summarization as a testbed.

\begin{figure}[t!]
        \centering
        \includegraphics[height=3.5in]{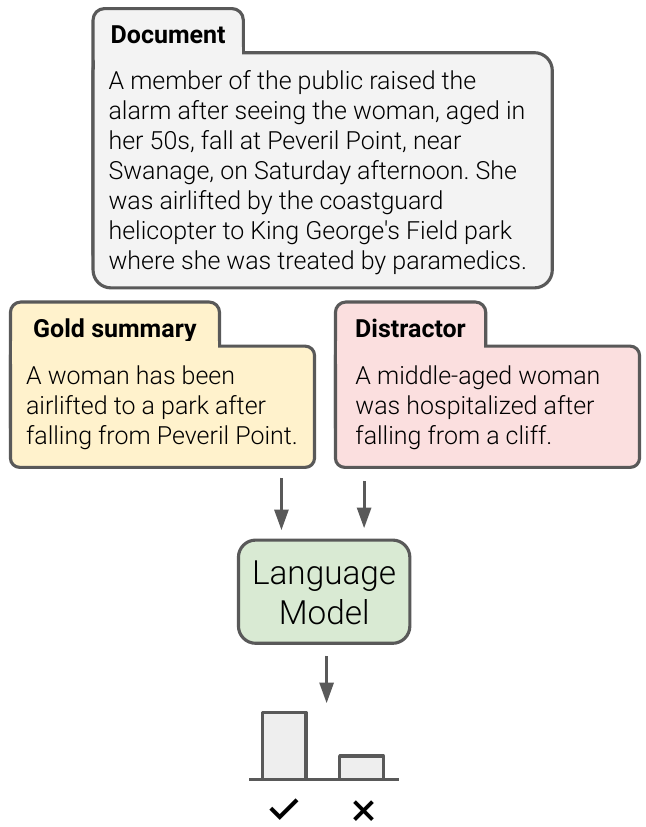}
        \caption{A schematic diagram of \benchmark, where we measure whether an LLM assigns a higher score to a \fc{} document summary than a  \fic{} summary. }
        \label{fig:intro}
\end{figure}

To achieve this goal, we propose \benchmark(the \textbf{F}actual \textbf{I}nconsistency \textbf{B}enchmark) to measure how often models prefer \fc{} summaries over \fic{} summaries. 
In \benchmark, models are given a document and are evaluated on whether they assign a higher score to a \fc{} summary than a \fic{} summary.
Scores are assigned based on a model's assigned probability to the summary.
We use accuracy on this binary classification task as a proxy for how factually consistent a model is.
\benchmark consists of over 3,500 pairs of summaries that were \emph{all} manually annotated as either \fc{} or \fic{}. 
The benchmark is based on documents and summaries from the XSum \citep{narayan2018don} and CNN/DM \citep{hermann2015teaching} datasets to test behavior on abstractive and extractive summarization, respectively. 
For \fc{} summaries, we use reference summaries from the datasets that we verify are \fc{} or manually edit to make them \fc{}.
The \fic{} summaries were generated from 22 models trained for summarization and then annotated as \fic{}.

To explore the behavior of existing models on \benchmark, we evaluate 23 LLMs from 6 different model families including BLOOM, OPT, GPT, and T0 ~\cite{radford2019gpt2, zhang2022opt, sanh2021multitask, chung2022scaling, lester-etal-2021-power, bloom2022} ranging from 1B to 176B parameters.
Next, we analyze whether the method used to generate the \fic{} summaries affects how often models prefers \fc{} summaries over \fic{} summaries.
To do so, we evaluate these models on \fic{} summaries from three additional sources: (1) unedited reference summaries that we annotated as factually inconsistent, (2) summaries edited via  FactCC~\cite{kryscinski-etal-2020-evaluating}, and (3) summaries produced by MFMA~\cite{lee-etal-2022-masked}. 
In addition, we test 4 different scoring functions: conditional log-likelihood (LL), length-normalized LL, pointwise mutual information (PMI), and length-normalized PMI.
Overall, we find that: 
(1) The LLMs we consider typically assign a higher score to factually consistent summaries than to factually inconsistent summaries (e.g.\ $72.4\%$ of the time for BLOOM \citep{bloom2022}), but 
(2) LLMs rarely prefer factually consistent summaries over factually inconsistent summaries copied verbatim from the document (e.g.\ $9.6\%$ of the time for BLOOM), 
(3) LLMs generally become more factually consistent as they are scaled up, and
(4) FactCC-generated factually inconsistent summaries can fool some LLMs at a similar rate to model-generated factually inconsistent summaries.

In summary, our contributions are: (1) a benchmarking procedure and collection of annotated summaries for probing the factual consistency of LLMs and (2) a thorough evaluation of 23 LLMs from 6 different model families of up to 176B parameters.
We hope \benchmark and our results help shed light on the factuality of LLMs.

\section{Related Work}

\subsection{Factuality Evaluation Datasets} 
In the literature on text summarization, many datasets with human-labeled factually consistent and inconsistent summaries have been introduced for meta-evaluation purposes (i.e., evaluating factuality evaluation metrics) or for training the metrics themselves.  \citet{pagnoni-etal-2021-understanding} introduced the FRANK benchmark that contains 2250 model-generated summaries with factuality labels for each summary sentence. Similarly, \citet{gabriel-etal-2021-go} proposed the GO FIGURE meta-evaluation framework that has 1500 model-generated summaries that include factuality labels. Besides these two benchmarks, many other works collected their own small-scale factuality evaluation datasets for evaluating their proposed metrics or analyzing the factuality of summarization models~\cite{falke-etal-2019-ranking, maynez-etal-2020-faithfulness, kryscinski-etal-2020-evaluating, wang-etal-2020-asking, durmus-etal-2020-feqa, lux-etal-2020-truth}. \citet{ribeiro-etal-2022-factgraph} combined labeled datasets from four works and formed the FactCollect dataset with more than 9000 summary sentences and their factuality labels. Additionally, a few other works proposed to automatically obtain \fic{} summaries by perturbing the reference summaries~\cite{kryscinski-etal-2020-evaluating, lee-etal-2022-masked}, e.g., entity swapping. However, \citet{goyal-durrett-2021-annotating} showed that these automatic techniques target inherently different error distributions than those seen in actual model generations.  \citet{goyal-durrett-2020-evaluating} considered model outputs at the top of beam search as factual and bottom generations as non-factual. The aforementioned works mainly focus on abstractive summarization; in contrast, \citet{zhang2022extractive} introduced a factuality evaluation dataset for extractive summarization which we use as part of \benchmark. 
Previous datasets do not annotate reference summaries and instead only annotate model generations as \fc{} or \fic{}.
However, the reference summaries are not always \fc{} \cite{maynez-etal-2020-faithfulness, bommasani-cardie-2020-intrinsic, tejaswin-etal-2021-well} which means that some of the \fic{} summaries might not have any \fc{} summary to pair with. Hence, we perform a manual verification of reference summaries as \fc{} for \benchmark.  
Additionally, \benchmark aims to evaluate the factual consistency of LLMs themselves instead of meta-evaluating evaluation metrics.

Besides summarization, \citet{devaraj-etal-2022-evaluating} proposed a factuality evaluation dataset for text simplification. In addition, some datasets have been introduced for checking a fact or claim against a large knowledge base~\cite{thorne-etal-2018-fever, augenstein-etal-2019-multifc}; here, we instead focus on factual consistency of conditional model continuations.

\subsection{Factuality Evaluation Metrics} 
Many metrics have been proposed to evaluate the factual consistency of model-generated summaries. These metrics can be roughly categorized into entailment-based metrics and question-generation/answering (QA/QG)-based metrics. Entailment-based metrics check whether each summary sentence (or a more fine-grained subsentence) is entailed by the source document~\cite{falke-etal-2019-ranking, kryscinski-etal-2020-evaluating, goyal-durrett-2020-evaluating, maynez-etal-2020-faithfulness}. QA/QG-based metrics are designed based on the idea that a question should have the same answer whether it is based on the summary or the document~\cite{wang-etal-2020-asking, durmus-etal-2020-feqa, scialom-etal-2021-questeval}. Relatedly, \citet{goodrich2019fac_acc} evaluated facutality by checking factual tuples extracted by OpenIE and \citet{ribeiro-etal-2022-factgraph} used the AMR graphs of the summary and the document for assessing factual consistency. All these metrics were designed to evaluate models trained specifically for summarization. In this work, we focus more broadly on evaluating the factual consistency of LLMs.     

\section{\benchmark: Factual Inconsistency Benchmark}

Each example in \benchmark consists of a document and two summaries: a \fc{} summary and a \fic{} summary. 
Models are evaluated based on the proportion of times they assign a higher score to a \fc{} summary than to a \fic{} summary. 
We define a \fc{} summary as a summary whose contents can be inferred solely from the document.
This means that even if a summary contains true information, if the information is not found in the document, then the summary is factually inconsistent. 
For example, the Gold summary in \cref{fig:intro} is factually consistent as it is written, but if we swapped \textit{Peveril Point} with \textit{a cliff}, then it would no longer be factually consistent, even if \textit{Peveril Point} is technically \textit{a cliff}, since this fact cannot be inferred from the document. 

We compare the factual consistency of models on both extractive and abstractive summaries.
Extractive summaries occur verbatim in the document while abstractive summaries do not. 
We use two summarization datasets as our testbed: CNN/DM \citep{see-etal-2017-get, hermann2015teaching} for extractive summaries and XSum \citep{narayan-etal-2018-dont} for abstractive summaries.
CNN/DM consists of English documents about the news from CNN/Daily Mail and summaries that are several sentences long with 287K/13K/11K examples for train/val/test.\footnote{\url{https://huggingface.co/datasets/cnn_dailymail}}
XSum consists of English documents about the news from BBC and short summaries with 204K/11K/11K examples for train/val/test.\footnote{\url{https://huggingface.co/datasets/xsum}}
The CNN/DM dataset is distributed under an Apache 2.0 license and 
 XSum is under a Creative Commons Attribution 4.0 International license. 
Our use is consistent with the intended use and we release our code under an Apache 2.0 license and the data for \benchmark under a Creative Commons Attribution 4.0 International license. 

\subsection{Dataset Construction}
We describe how we construct the \fc{} and \fic{} summaries for \benchmark.
When performing annotations, each summary was annotated by two annotators. 
Four of the authors performed the annotations.
Our inter-annotator agreement was $91.3\%$.
Whenever there was a disagreement on a given summary, the two annotators would discuss and resolve the disagreement. 
See \cref{sec:annotation_instructions} for annotator instructions.

\paragraph{Factually Consistent Summaries.}
Though the summarization datasets we consider include reference summaries, the reference summaries are not necessarily factually consistent with the document~\cite{maynez-etal-2020-faithfulness}. 
To account for this, we annotate reference summaries for 500 and 100 documents from XSum and CNN/DM respectively as either \fc{} or \fic{}.
Then, we edit the \fic{} reference summaries to be \fc{} using minimal edits. 
Factually inconsistent reference summaries usually contain information that is true but not found in the document.
Thus, most edits involve removing or changing certain keywords or phrases not present in the document. 
Two annotators then verified the edited summary was \fc{}.
The percentage of \fc{} summaries that were edited from the original reference summary was roughly $90\%$ for XSum and $30\%$ for CNN/DM.
We denote these annotated \fc{} reference summaries as \emph{Gold} summaries.
See \cref{sec:sample_edited_summaries} for some examples of edited summaries.

\paragraph{Factually Inconsistent Summaries.}
To obtain \fic{} summaries, we generate summaries from models trained on a given summarization dataset and annotate the generated summaries as \fc{} or \fic{}.
We then retain the model-generated summaries that were annotated as \fic{}. 
We use 15 extractive models to generate summaries for CNN/DM and 7 generative models to generate summaries for XSum. 
See \cref{sec:models_to_generate_summaries} for the list of models used to generate the summaries.
For XSum, we annotate the model-generated summaries ourselves and for CNN/DM we source the factual-consistency annotations from \citet{zhang2022extractive}. 
See \cref{sec:sample_model_extracted_fic} for some examples of factually inconsistent model-extracted summaries.

For the dataset underlying our benchmark, we create a paired example for every possible factually inconsistent summary with the Gold summary for a given document.
In the end, we have 3,124 factually consistent/inconsistent summary pairs across 500 unique documents for XSum and 457 pairs across 96 unique documents for CNN/DM (4 CNN/DM documents were dropped since all the models generated factually consistent summaries for them).
A model's accuracy on \benchmark is then simply the proportion of summary pairs where the model assigns a higher score to the Gold summary than to the \fic{} summary.

\subsection{Scoring Function}

For \benchmark, we are primarily interested in a scoring function to measure the consistency of the summary and the document.
A natural scoring function is the model's assigned log-likelihood (LL) of the summary given the document, but LL has two major issues.
First, the log-likelihood has a bias towards shorter summaries since the probability of each token in a summary is multiplied together to obtain the log-likelihood of the entire summary, and thus shorter summaries tend to produce higher log-likehoods. 
Second, if the summary alone has a high likelihood, then the model might assign a high likelihood to the summary, even if the summary and the document are not that related. 
To address the first issue, we normalize by the length of the summary.
To address the second issue, we use the pointwise mutual information (PMI), which accounts for the likelihood of the summary by subtracting the log-likelihood of the summary alone from the log-likelihood of the summary conditioned on the document.
Several recent works have used the pointwise mutual information (PMI) as a way of scoring a language model's generations:
\citet{holtzman-etal-2021-surface} used PMI to solve multiple-choice tasks that probe for knowledge using GPT3 and \citet{padmakumar-he-2021-unsupervised} used PMI for unsupervised extractive summarization.
Concurrently, \citet{van2022mutual} show that optimizing for PMI during decoding can decrease hallucinations in language models. 

To address both these issues, we use the length-normalized PMI as our default scoring function, where the length normalization is performed by averaging over tokens.
Specifically, given document $d$ and summary $s$ which consists of $T$ tokens $\{ s_{1}, s_{2},  ... , s_{T} \}$, the length-normalized PMI is defined as
\begin{align} \label{eq:pmi}
 \frac{1}{T} \log \sum_{t=1}^T P(s_{t}|d, s_{1}, ..., s_{t-1}) \\
 - \frac{1}{T} \log \sum_{t=1}^T P(s_{t}|, s_{1}, ..., s_{t-1}) \nonumber
\end{align}
We ablate the impact of using different scoring functions in \cref{sec:scoring}.

\section{Experiments}
Having defined our benchmark, we now evaluate the factual consistency of various LLMs and compare with several other methods for generating alternative summaries and assigning scores to LM generations.

\subsection{Models}

We evaluate 23 large language models (1B to 176B parameters) from 6 different model families: 
\begin{itemize}[leftmargin=*]
    \item \textbf{GPT:} GPT2-XL \citep{radford2019gpt2}, GPT-Neo-1.3B, GPT-Neo-2.7B, GPT-NeoX-20B \citep{black-etal-2022-gpt}
    \item \textbf{OPT:} OPT-1.3B, OPT-2.7B, OPT-6.7B, OPT-13B, OPT-30B, OPT-66B, OPT-175B \cite{zhang2022opt}
    \item \textbf{BLOOM:} BLOOM-1.1B, BLOOM-1.7B, BLOOM-3B, BLOOM-7B, BLOOM~\cite{bloom2022}
    \item \textbf{T0:} T0-3B, T0 \cite{sanh2021multitask}
    \item \textbf{FLAN-T5:} FLAN-T5-XL, FLAN-T5-XXL \citep{chung2022scaling}
    \item \textbf{T5-LM-Adapt:} T5-LM-Adapt-XL, T5-LM-Adapt-XXL \citep{lester-etal-2021-power}
\end{itemize}
Our chosen models consist of both zero-shot models that were not trained on XSum or CNN/DM (GPT, OPT, BLOOM, T5-LM-Adapt) and models that were trained on XSum and CNN/DM in a multi-task fashion (T0, FLAN-T5). 
For each model, we use the same 3 prompts and report the median performance across prompts, following \citet{sanh2021multitask}.
See \cref{sec:prompt_templates} for the prompt templates used.
We use a maximum sequence length of 512, which was also applied when sampling 500 documents from XSUM for annotating factual consistency.
We use Pytorch \cite{paszke2019pytorch} and HuggingFace \cite{wolf-etal-2020-transformers} to run the models, and use bitsandbytes \cite{dettmers2022llmint8} to do 8-bit inference for the larger models.
All experiments were run on NVIDIA A6000s or 80GB NVIDIA A100s (depending on the model) and took about two days.

\begin{figure}[t]
 
    \centering
    \includegraphics[width=\columnwidth]{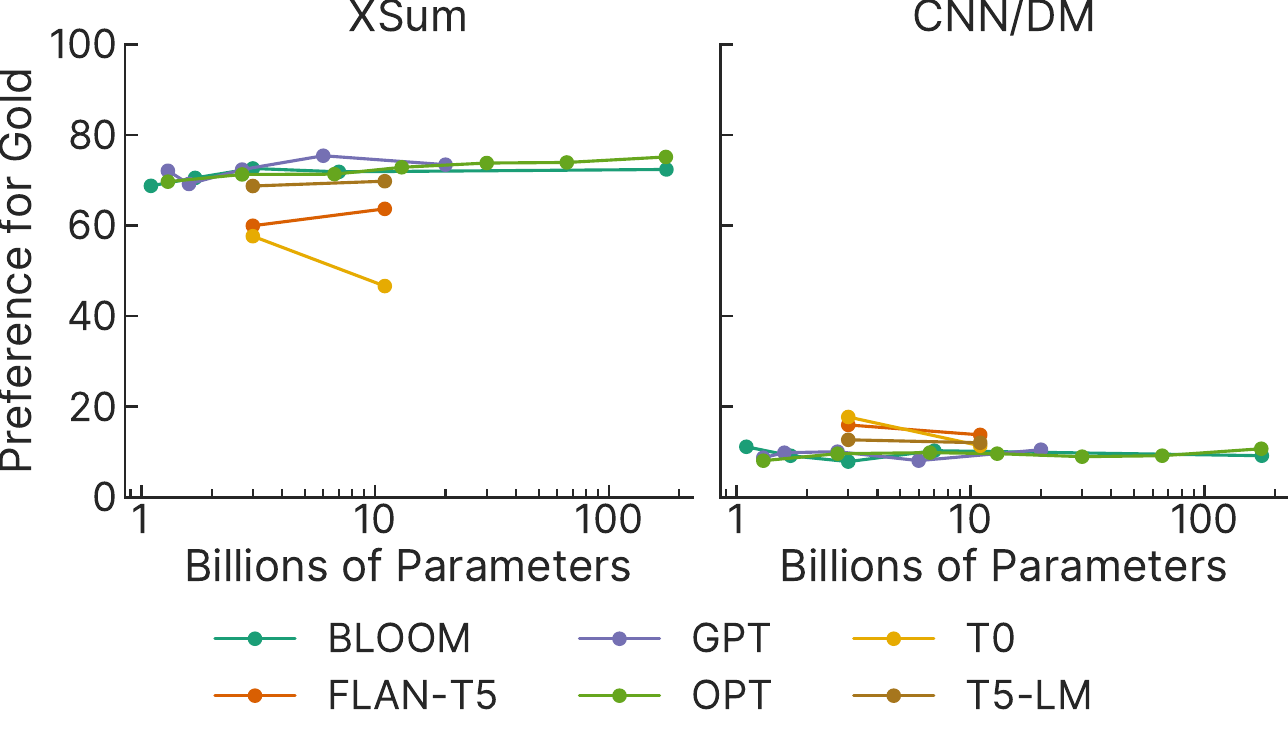}
    \caption{Performance of various models on \benchmark.}
    \label{fig:fib}    

\end{figure}

\subsection{Main Results}

We show the performance of all the models on XSum and CNN/DM in \cref{fig:fib}.
On XSum, we highlight the following: 
\begin{itemize}[leftmargin=*]
    \item \textit{Factual Consistency:} Models generally prefer Gold summaries over \fic{} model-generated summaries, but the average accuracy of any model is still far from 100\%.
    \item \textit{Effect of Scale:} Performance generally increases slightly with scale within a given model family with the exception of T0, where the 11-billion-parameter model underperforms T0-3B. For zero-shot LLMs, the performance is remarkably similar across model families.
    \item \textit{Effect of Training:} Both FLAN-T5 and T0 underperform the zero-shot models, which could be because they were trained on the XSum dataset, which had many reference summaries that were factually inconsistent.
\end{itemize}

In contrast to our results on XSum, we find that models rarely assign a higher score to \fc{} reference summaries than to \fic{} model-extracted summaries on the CNN/DM dataset.
However, if the \fc{} summary is also model-extracted, then models also assign higher scores to the \fc{} model-extracted summary. 
This suggests that all models have a strong preference for text copied from the input regardless of its factual-consistency.

\begin{figure*}[ht!]
 
    \centering
    \includegraphics[width=0.9\textwidth]{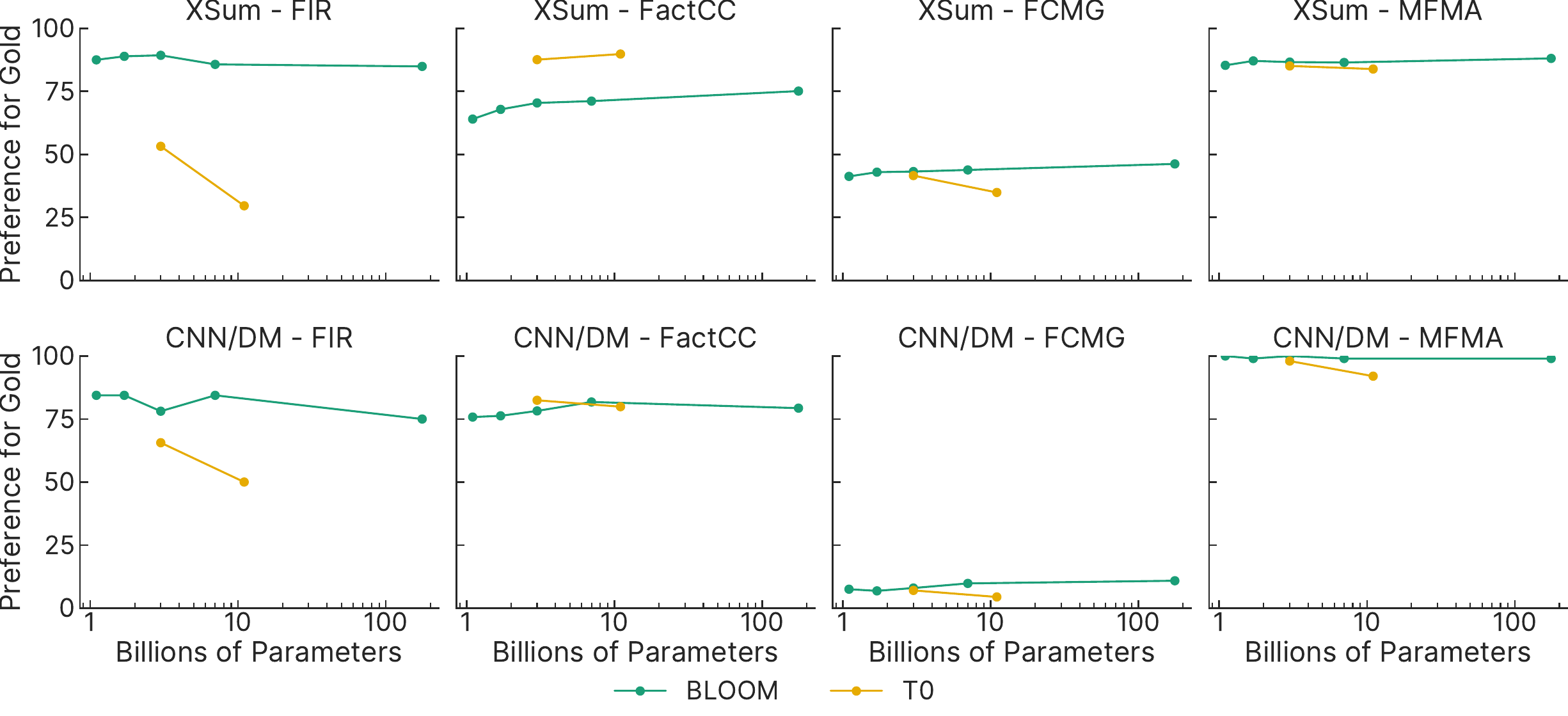}
    \caption{Preference for the Gold summary exhibited by BLOOM and T0 when using different methods for generating alternative choices.}
    \label{fig:alternative}    

\end{figure*}

\subsection{Generating Alternative Summaries}
We also analyze the impact of the the method used to generate \fic{} summaries.
To do so, we compare the model's performance when using different methods for generating the \fic{} summary.
We note that \citet{goyal-durrett-2021-annotating} showed that these automatic techniques target inherently different error distributions than those seen in actual model generations. 
We experiment with the following alternative methods for obtaining \fic{} summaries: 
\begin{itemize}[leftmargin=*]
    \item MFMA, proposed by \citet{lee-etal-2022-masked}, uses pre-trained masked language models to generate \fic{} summaries.
    Specifically, summaries are generated by reconstructing the reference summary conditioned on the document and reference summary with $\alpha$ and $\beta$ percent of the entities masked out respectively.
    The MFMA procedure first fine-tunes a pre-trained masked LM to reconstruct summaries in this setup and then uses the fine-tuned model to generate new summaries.
    For example, in \cref{fig:intro}, if we masked out \textit{Peveril Point} in the reference summary and the model generated \textit{the grand canyon} instead, then the factually-inconsistent MFMA-generated summary would be \textit{A middle-aged woman has been driven by ambulance to a park after falling from the grand canyon.}
    We follow the setup in MFMA and use T5-base \citep{raffel2020exploring} and BART-base \citep{lewis2020bart} to generate the summaries with $\alpha = 0.8$ and $\beta = 0.6$.
    Since there is no guarantee that the model-reconstructed summaries are \fic{}, we annotate their factual-consistency and only keep the ones that are \fic{}.
    We construct \fic{} summaries from MFMA by combining all \fic{} summaries generated by T5-base and BART-base.
    \item FactCC, proposed by \citet{kryscinski-etal-2020-evaluating}, generates \fic{} summaries via heuristic perturbations to reference summaries.
    FactCC uses two ways to perturb the reference summary: entity swapping and sentence negation.
    Entity swapping replaces an entity (i.e.\ pronouns, dates, numbers and named entities) in the reference summary with a different entity from the document and sentence negation refers to negating a verb.
    For example, in \cref{fig:intro}, if we negated \textit{has} to \textit{hasn't}, then the factually-inconsistent FactCC-generated summary would be \textit{A middle-aged woman hasn't been airlifted to a park after falling from Peveril Point.} 
    \item FIR (\textbf{f}actually \textbf{i}nconsistent \textbf{r}eference) summaries. Since some of the original reference summaries were \fic{} and had to be edited to become \fc{}, we use these original reference summaries as an alternative source of \fic{} summaries.
\end{itemize}

As an additional baseline, we consider using \fc{} model-generated summaries rather than a \fic{} summary as the alternative summary.
This allows us to test whether models prefer model-generated summaries over Gold summaries.
We call this setup of where the alternative choice is a \fc{} model-generated summaries FCMG (\textbf{F}actually-\textbf{C}onsistent \textbf{M}odel-\textbf{G}enerated summaries).

A comparison of different methods for generating alternative summaries is shown in \cref{fig:alternative}.
We only plot results for BLOOM and T0 since the results for other decoder-only zero-shot LLMs are similar to those for BLOOM and the results for FLAN-T5 are similar to T0.
We highlight the following trends:

\begin{itemize}[leftmargin=*]
    \item \textit{Preference for \fc{} model-generated summaries depends on whether summaries are extractive:} On XSum, models are almost at chance when distinguishing between \fc{} model-generated summaries and Gold summaries.
    This is evident from the accuracy on FCMG being around $50\%$.
    However, on CNN/DM, models consistently prefer \fc{} model-extracted summaries to Gold summaries.
    We conclude that models prefer model-extracted summaries that occur verbatim in the document, regardless of their factual consistency. 
    \item \textit{MFMA's Ineffectiveness:} On both XSum and CNN/DM, models rarely assign MFMA-generated summaries a higher score than Gold summaries -- the accuracy on MFMA is between $85\%$ to $100\%$ across all models.
    \item \textit{FactCC's Effectiveness for zero-shot LLMs:}
    On XSum, BLOOM's performance is similar when either FactCC or model-generated \fic{} summaries are used as an alternative, and on CNN/DM, performance is similar for FactCC and \fic{} reference summaries.
    This suggests that FactCC generates somewhat plausible \fic{} summaries for  zero-shot decoder-only LLMs.
    \item \textit{FactCC's Effectiveness for other models:} However, T0, FLAN-T5, and T5-LM-Adapt (see \cref{sec:factcc_scores} for FLAN-T5 and T5-LM-Adapt accuracies) all perform better when using FactCC-generated \fic{} summaries than when using model-generated \fic{} summaries. 
    This indicates FactCC might not be effective in generating plausible \fic{} summaries across all model architectures and training schemes.
    \item \textit{Preference for Edited Summaries:} On XSum and CNN/DM, models tend to prefer \fc{} reference summaries over \fic{} reference summaries. 
    This is evident from the accuracy on FIR being around $80\%$ and indicates that models tend to prefer factually consistent summaries over factually inconsistent summaries.   
\end{itemize}

\begin{figure*}[ht!]
 
    \centering
    \includegraphics[width=0.9\textwidth]{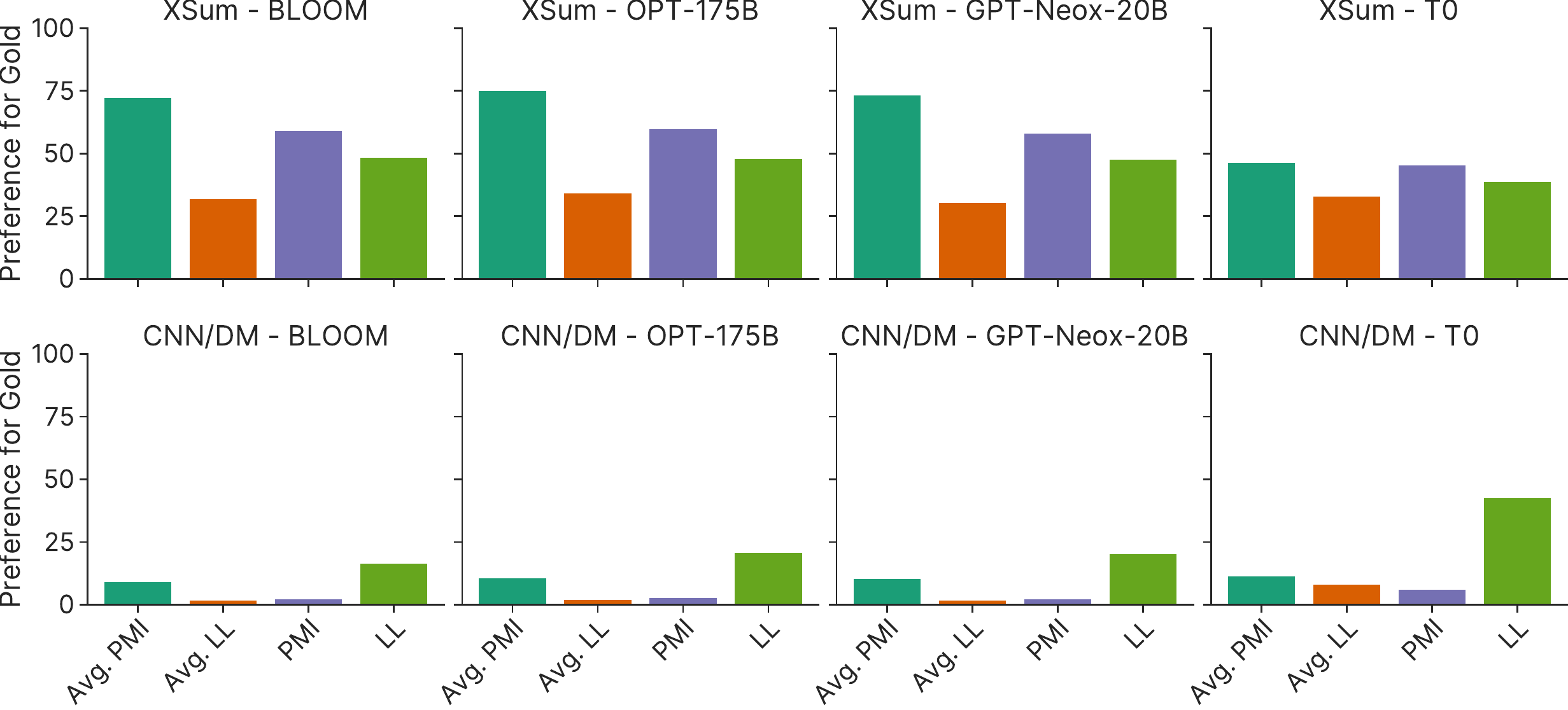}
    \caption{Performance of various models on \benchmark when using different scoring functions.}
    \label{fig:scoring}    

\end{figure*}

\subsection{Scoring Function}
\label{sec:scoring}

In \benchmark, we use the length-normalized PMI as the scoring function.
To validate this choice, we compare various alternative scoring functions: standard log-likelihood, length-normalized log-likelihood, and the non-length-normalized PMI.
We show results for BLOOM, OPT-175B and T0 on XSum and CNN/DM using different scoring methods in \cref{fig:scoring}.
In general we see that the average PMI enables models to best distinguish between \fc{} and \fic{} summaries. 
We also compare each scoring function on the alternate sources of \fic{} summaries; see \cref{sec:total_scores} for detailed results.
We find that log-likelihood works best when the factually inconsistent summary was produced by FactCC or is a model generation on CNN/DM.
We hypothesize that log-likelihood works better than length-normalized PMI on FactCC because the generated summaries are often non-fluent and therefore are assigned a low likelihood regardless of their factual consistency. 
For model-extracted summaries on CNN/DM, we hypothesize that log-likelihood works better than length-normalized PMI because log-likelihood is not as biased towards summaries extracted from the document as PMI is. 

\section{Analysis}

\begin{table*}
\centering
\footnotesize
\begin{tabular}{P{6.5cm}|P{4cm}|P{4cm}}
\toprule
    \textbf{Document} & \textbf{Factually Consistent Summary} & \textbf{Factually Inconsistent Summary} \\
\midrule
    The \$5m (3.2m) prize is supposed to be awarded each year to an elected leader who governed well, raised living standards and then left office. This is the fourth time in five years there has been no winner ... Sudan-born telecoms entrepreneur Mr Ibrahim launched the prize in an attempt to encourage African leaders to leave power peacefully. ... & The prize from Ibrahim for good governance in Africa has gone unclaimed yet again. & The winner of the prestigious Africa Leadership Prize has been announced by the African Union's executive committee. \\
\midrule
    The character with a huge papier mache head ... Hundreds of people attended an unveiling ceremony earlier, many in fancy dress for the occasion. Neil Taylor, who helped raise the donations for the statue, said its installation would mean that \"Frank will gaze on the Timperley sunset forever\" ... Frank Sidebottom created a whole ... & A statue of the character Frank Sidebottom has been unveiled in Timperley. & A statue of Timperley's character Frank Sidebottom has been unveiled at a Manchester museum.\\
\bottomrule
    \end{tabular}
    \caption{Two examples where BLOOM assigns a higher score to the \fic{} model-generated summaries than the Gold summary. These examples have id 24521870 and id 24601038 respectively.}
    \label{tab:examples}
\end{table*}

To get a better sense of what kind of factually inconsistent model-generated summaries tend to fool models into assigning a higher score than the Gold summary, we show some examples for BLOOM in \cref{tab:examples}.
These \fic{} summaries consist of extrinsic hallucinations that add new information rather than intrinsic hallucinations that manipulate the information in the document \citep{maynez-etal-2020-faithfulness}. 
In addition, these \fic{} summaries contain information that is actually false, not just information absent from the document.

\subsection{Factual Consistency of Models Used to Generate Summaries}
\begin{figure}[t]
    \centering
    \includegraphics[height=2.3in]{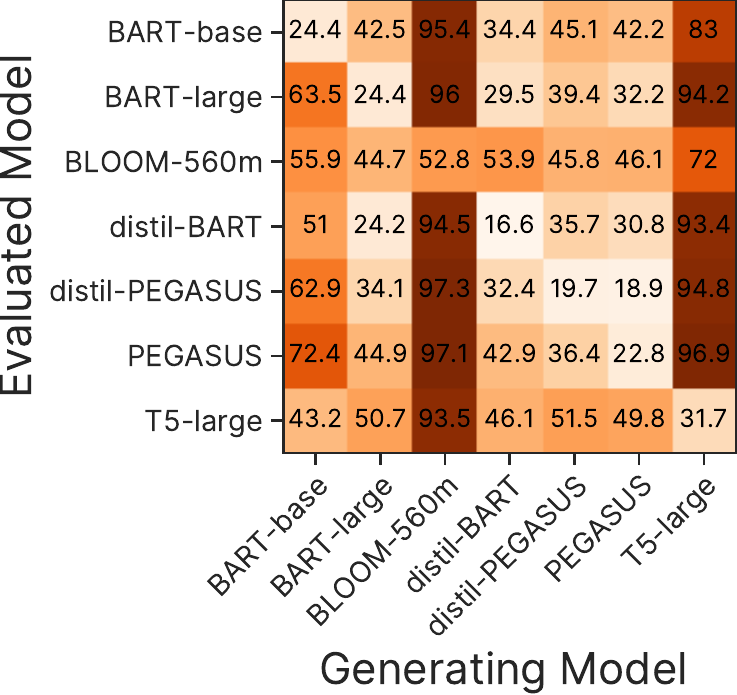}
    \caption{Heatmap showing the rate at which an ``evaluated  model`` assigns a Gold summary on XSum a higher score than a \fic{} summary generated by the ``generating model``. 
    }
    \label{fig:heatmap}
\end{figure}

We take the models used to generate the factually inconsistent summaries for XSum and evaluate them against each other using the same procedure as in \benchmark.
Specifically, we use \fic{} summaries produced by a ``generating model'' and measure how often an ``evaluated model'' assigns a higher score to the Gold summary than it does to the \fic{} model-generated summaries.
The result is summarized in \cref{fig:heatmap}, with full results in \cref{sec:model_trained}.
The accuracies down the diagonal are the lowest, which means models perform poorly when scoring their own \fic{} summary. 
This is expected since models should give high scores to \fic{} summaries they generate.
In most cases, Gold summaries are preferred less than 50\% of the time, suggesting that summarization models tend to assign higher scores to model-generated \fic{} summaries.
However, certain models (BLOOM and T5-large) almost always produce summaries that are assigned low scores by the other models.
We leave exploration of this trend to future work.

\section{Conclusion and Takeaways}
We present \benchmark, a new benchmark for evaluating the factual consistency of language models, and evaluate 23 large language models on \benchmark.
Our takeaways are: (1) LLMs tend to assign higher scores to \fc{} summaries than to \fic{} summaries, except that LLMs almost always assign higher scores to extracted summaries even if they are \fic{} and 
(2) length-normalized PMI enables models to most effectively detect \fic{} summaries.
Our results open new avenues for future work, including a more fine-grained study on the type of \fic{} errors different LLMs make and investigating the effect training on summarization has on the factual consistency of LLMs.

\section{Limitations}

One limitation with \benchmark is that it only measures the factual consistency of language models for the task of summarization, and specifically news summarization. 
It is not clear how well the results will generalize, for example, to other domains such as scientific article or other tasks such as question answering. 

\section*{Acknowledgements}

This work was supported by NSF-AI Engage Institute DRL-2112635.

\bibliography{anthology,custom}
\bibliographystyle{acl_natbib}

\appendix

\section{Annotation Instructions}
\label{sec:annotation_instructions}

The annotators were instructed to mark a summary as factually inconsistent if any information in the summary was not implied in the document. We assume no access to external knowledge so the summary has to be implied solely from the document. External knowledge is broadly defined as any knowledge that cannot be inferred from common sense alone. For example, the capital of a country or the rules of a sport would be external knowledge.

\section{Sample Edited Summaries}
\label{sec:sample_edited_summaries}

We show some examples of documents with the original \fic{} reference summary and the edited \fc{} summary on XSum in \cref{tab:edited_examples}.

\begin{table*}
\centering
\footnotesize
\begin{tabular}{P{6.5cm}|P{4cm}|P{4cm}}
\toprule
        \textbf{Document} & \textbf{Original Ref. Summary} & \textbf{Edited Ref. Summary} \\
\midrule
    West Midlands Ambulance Service said the car was discovered on Sunday at 09:35 GMT by two cyclists in Crakemarsh near Uttoxeter, Staffordshire. A spokesman said the black Ford Fiesta appeared to have hit a tree in very foggy conditions on the B5030. The girl, in the back of the car, was treated at hospital for minor injuries. The man, who was 25 and from the local area, has not yet been named ... & A five-year-old girl has been found with her dead father in a crashed car which had been in a ditch ``for some time''. & A girl has been found in a crashed car. \\
\midrule
    Aiden Webb, 22, from Norwich, was climbing Fansipan mountain alone on Friday when he fell down a ravine and lost his way ... in the fall on the 3,100m (10,300ft) high Fansipan mountain in the north of Vietnam ... A Foreign and Commonwealth Office spokeswoman said: "We are supporting the family of Aiden Webb, a British man reported missing in Vietnam. We are working closely with the local authorities leading the search." & A British man is missing in Vietnam after falling while attempting to climb the country's highest mountain. & A British man is missing in Vietnam after falling while attempting to climb a mountain.\\
\bottomrule
    \end{tabular}
    \caption{These examples have id 34696511 and id 36459564 respectively.}
    \label{tab:edited_examples}
\end{table*}

\section{Sample Model-Extracted \fic}
\label{sec:sample_model_extracted_fic}

We show some examples of documents with model-extracted factually inconsistent summaries on CNN/DM in \cref{tab:model_extracted_fic}.

\begin{table*}
\centering
\footnotesize
\begin{tabular}{P{8cm}|P{6cm}}
\toprule
    \textbf{Document} & \textbf{Model-Extracted Factually Inconsistent Summary} \\
\midrule
    the california public utilities commission on thursday said it is ordering pacific gas \& electric co. to pay a record 1.6 billion penalty ... 850 million will go to `` gas transmission pipeline safety infrastructure improvements , '' the commission said ... pg \& e failed to uphold the public 's trust , '' commission president michael picker said ... the company 's chief executive officer said ... `` since the 2010 explosion of our natural gas transmission pipeline in san bruno , we have worked hard to do the right thing for the victims , their families and the community of san bruno , '' tony earley said ... & 
        ... 850 million will go to `` gas transmission pipeline safety infrastructure improvements , '' the commission said .  `` since the 2010 explosion of our natural gas transmission pipeline in san bruno , we have worked hard to do the right thing for the victims , their families and the community of san bruno ... \\
\midrule
a passenger on an atlanta-bound air canada flight told a cnn reporter on the plane friday that a stranger sitting behind him tried to choke him . oliver minatel , 22 , said he was sleeping on air canada flight 8623 from toronto when he felt something around his neck  ... `` i forced it ( the cord ) down and then other people came to help , and then i got out and he started saying that we were here to kill him , '' minatel said . the man was not restrained for the rest of the trip , but the flight crew told him to stay seated with his seat belt on . the man kept trying to get out of his seat but other passengers yelled at him whenever he tried to stand up .
& oliver minatel , 22 , said he was sleeping on air canada flight 8623 from toronto when he felt something around his neck . the man kept trying to get out of his seat but other passengers yelled at him whenever he tried to stand up . the suspect was escorted off the plane . \\ 
\bottomrule
    \end{tabular}
    \caption{Two examples of model-extracted factually inconsistent summaries. The annotations were sourced from \citet{zhang2022extractive}. These examples have id 41c6edecee127c396d17e2e9115a4a89252cc52b and id 32655a04c9e4733a1ae4b210a045bc6e0d443d85 respectively. The first example uses Textrank \cite{mihalcea-tarau-2004-textrank} to extract the summary. It is factually incorrect since 'we' refers to pg \& e and not the commission. The second example uses MatchSumm \cite{zhong-etal-2020-extractive} to extract the summary. It is factually inconsistent since the man refers to the stranger and not Oliver Minatel.}
      \label{tab:model_extracted_fic}
\end{table*}

\section{Models Used to Generate Summaries} 
\label{sec:models_to_generate_summaries}
We use the following models to generate summaries for XSum and include the respective HuggingFace model name:
\begin{itemize}
    \item BLOOM-560m \cite{bloom2022} \textbf{-} mrm8488/bloom-560m-finetuned-news-summarization-xsum
    \item BART-base \cite{lewis-etal-2020-bart} \textbf{-} VictorSanh/bart-base-finetuned-xsum
    \item distil-PEGASUS \cite{zhang2020pegasus} \textbf{-} sshleifer/distill-pegasus-xsum-16-8 
    \item BART-large  \cite{lewis-etal-2020-bart} \textbf{-} facebook/bart-large-xsum
    \item PEGASUS \cite{zhang2020pegasus} \textbf{-} google/pegasus-xsum
    \item distil-BART  \cite{lewis-etal-2020-bart}  \textbf{-} sshleifer/distilbart-xsum-12-6
    \item T5-large \cite{raffel2020exploring}\textbf{-} sysresearch101/t5-large-finetuned-xsum
\end{itemize}
We use greedy decoding for all models with a maximum generation length of 50 tokens.

We use the following models to generate summaries for CNN/DM. See \citet{zhang2022extractive} for more description of the models.
\begin{itemize}
    \item Oracle \cite{lin-2004-rouge}
    \item Oracle (discourse) \cite{xu-etal-2020-discourse}
    \item RNN Ext RL \cite{chen-bansal-2018-fast}
    \item BanditSumm \cite{dong-etal-2018-banditsum}
    \item NeuSumm  \cite{zhou-etal-2018-neural-document}
    \item Refresh \cite{narayan-etal-2018-ranking}
    \item BERT+LSTM+PN+RL \cite{zhong-etal-2019-searching}
    \item MatchSumm \cite{zhong-etal-2020-extractive}
    \item HeterGraph \cite{wang-etal-2020-heterogeneous}
    \item Lead3 
    \item Textrank \cite{mihalcea-tarau-2004-textrank}
    \item Textrank (ST) \cite{reimers-gurevych-2019-sentence}
    \item PacSum (tfidf) \cite{zheng-lapata-2019-sentence}
    \item PacSum (bert)
    \item MI-unsup \cite{padmakumar-he-2021-unsupervised}

\end{itemize}

\section{Prompt Templates}
\label{sec:prompt_templates}
We use the following 3 prompt templates for all models, where [input] is replaced with the document:
\begin{itemize}
    \item "[input]"
    \item "The summary of "[input]" is "
    \item "Summarize: [input]"
\end{itemize}

\section{Accuracies Across All Scoring Functions}
\label{sec:total_scores}
We show the performance of all the models across different scoring functions for XSum in \cref{tab:xsum_total_avg_pmi}, \cref{tab:xsum_total_avg_ll}, \cref{tab:xsum_total_pmi}, and \cref{tab:xsum_total_ll} and for CNN/DM in \cref{tab:cnn_dm_total_avg_pmi}, \cref{tab:cnn_dm_total_avg_ll}, \cref{tab:cnn_dm_total_pmi}, and \cref{tab:cnn_dm_total_ll}.

\begin{table*}[!ht]
    \small
    \centering
    \begin{tabular}{@{} l c c c c c @{}}
        \toprule
        Model & FIR & FCMG & FIB & FactCC & MFMA \\
       \midrule
        T0-3B & 53.2 & 41.6 & 57.6 & 87.6 & 85.1 \\
        T0 & 29.6 & 34.9 & 46.6 & 89.8 & 83.9 \\
        FLAN-T5-xl & 58.1 & 47.8 & 59.9 & 87.3 & 85.6 \\
        FLAN-T5-xxl & 59.0 & 51.3 & 63.7 & 87.1 & 87.3 \\
        T5-LM-Adapt-xl & 81.3 & 49.5 & 68.7 & 78.7 & 87.5 \\
        T5-LM-Adapt-xxl & 81.7 & 50.7 & 69.8 & 84.2 & 88.7 \\
        GPT-Neo-1.3B & 88.0 & 45.7 & 72.1 & 68.9 & 87.1 \\
        GPT2-XL & 84.9 & 46.3 & 69.2 & 71.5 & 83.2 \\
        GPT-Neo-2.7B & 87.8 & 47.7 & 72.3 & 72.2 & 85.1 \\
        GPTJ-6B & 88.0 & 51.2 & 75.4 & 74.0 & 87.3 \\
        GPT-Neox-20B & 82.9 & 49.6 & 73.4 & 74.1 & 86.4 \\
        BLOOM & 84.9 & 46.2 & 72.4 & 75.1 & 88.1 \\
        BLOOM-7B1 & 85.7 & 43.8 & 71.8 & 71.1 & 86.5 \\
        BLOOM-3B & 89.3 & 43.2 & 72.6 & 70.4 & 86.6 \\
        BLOOM-1B7 & 88.9 & 42.9 & 70.5 & 67.8 & 87.1 \\
        BLOOM-1B1 & 87.5 & 41.3 & 68.8 & 64.0 & 85.3 \\
        OPT-175B & 84.4 & 48.3 & 75.1 & 71.2 & 87.0 \\
        OPT-66B & 83.5 & 47.8 & 73.9 & 70.8 & 87.2 \\
        OPT-30B & 84.4 & 48.3 & 73.8 & 72.0 & 87.2 \\
        OPT-13B & 85.1 & 49.0 & 72.9 & 71.6 & 86.5 \\
        OPT-6.7B & 83.3 & 47.4 & 71.3 & 70.5 & 86.3 \\
        OPT-2.7B & 84.4 & 48.1 & 71.3 & 70.5 & 85.8 \\
        OPT-1.3B & 85.7 & 46.3 & 69.7 & 70.5 & 86.0 \\
        \bottomrule
    \end{tabular}
    \caption{The performance of the models on XSum with various alternative-choices using avg. PMI as the scoring function.}
    \label{tab:xsum_total_avg_pmi}
\end{table*}

\begin{table*}[!ht]
    \small
    \centering
    \begin{tabular}{@{} l c c c c c @{}}
        \toprule
        Model & FIR & FCMG & FIB & FactCC & MFMA \\
       \midrule
        T0-3B & 20.0 & 15.5 & 29.1 & 97.7 & 68.2 \\
        T0 & 14.9 & 21.4 & 33.0 & 96.9 & 73.2 \\
        FLAN-T5-xl & 23.6 & 16.2 & 29.4 & 97.7 & 68.9 \\
        FLAN-T5-xxl & 21.6 & 17.6 & 32.1 & 98.1 & 72.0 \\
        T5-LM-Adapt-xl & 34.1 & 17.7 & 23.9 & 93.1 & 62.3 \\
        T5-LM-Adapt-xxl & 28.1 & 19.2 & 26.4 & 95.7 & 67.0 \\
        GPT-Neo-1.3B & 37.4 & 18.1 & 24.7 & 94.7 & 59.1 \\
        GPT2-XL & 33.6 & 19.3 & 26.0 & 95.3 & 60.7 \\
        GPT-Neo-2.7B & 35.9 & 19.5 & 26.9 & 95.8 & 62.0 \\
        GPTJ-6B & 28.3 & 21.1 & 28.4 & 96.8 & 68.9 \\
        GPT-Neox-20B & 23.4 & 20.8 & 30.5 & 97.0 & 69.8 \\
        BLOOM & 26.5 & 24.3 & 32.1 & 97.8 & 73.1 \\
        BLOOM-7B1 & 39.9 & 21.5 & 28.8 & 96.3 & 65.6 \\
        BLOOM-3B & 44.3 & 20.5 & 28.2 & 95.7 & 63.9 \\
        BLOOM-1B7 & 49.0 & 20.8 & 27.1 & 94.7 & 61.2 \\
        BLOOM-1B1 & 51.4 & 20.4 & 27.4 & 93.0 & 59.7 \\
        OPT-175B & 16.9 & 23.1 & 34.4 & 97.9 & 77.1 \\
        OPT-66B & 18.7 & 22.8 & 32.3 & 97.5 & 75.1 \\
        OPT-30B & 20.3 & 21.6 & 32.6 & 97.4 & 72.4 \\
        OPT-13B & 22.5 & 21.4 & 31.0 & 96.6 & 73.2 \\
        OPT-6.7B & 22.0 & 21.3 & 28.7 & 96.7 & 70.2 \\
        OPT-2.7B & 29.0 & 20.1 & 28.4 & 96.7 & 68.7 \\
        OPT-1.3B & 30.7 & 19.9 & 26.3 & 95.9 & 64.7 \\
        \bottomrule
    \end{tabular}
    \caption{The performance of the models on XSum with various alternative-choices using avg. LL as the scoring function.}
    \label{tab:xsum_total_avg_ll}
\end{table*}

\begin{table*}[!ht]
    \small
    \centering
    \begin{tabular}{@{} l c c c c c @{}}
        \toprule
        Model & FIR & FCMG & FIB & FactCC & MFMA \\
       \midrule
        T0-3B & 18.3 & 46.0 & 49.1 & 83.2 & 83.7 \\
        T0 & 16.7 & 36.8 & 45.6 & 89.0 & 83.7 \\
        FLAN-T5-xl & 16.7 & 52.0 & 49.0 & 82.0 & 82.9 \\
        FLAN-T5-xxl & 16.7 & 51.2 & 53.6 & 81.3 & 85.6 \\
        T5-LM-Adapt-xl & 39.0 & 52.6 & 54.7 & 69.9 & 83.8 \\
        T5-LM-Adapt-xxl & 35.4 & 51.5 & 55.3 & 76.8 & 85.1 \\
        GPT-Neo-1.3B & 58.4 & 46.5 & 57.2 & 60.5 & 83.9 \\
        GPT2-XL & 56.1 & 51.6 & 54.9 & 64.5 & 80.2 \\
        GPT-Neo-2.7B & 57.5 & 49.4 & 55.2 & 66.3 & 82.3 \\
        GPTJ-6B & 55.7 & 54.9 & 57.8 & 66.7 & 84.3 \\
        GPT-Neox-20B & 53.0 & 49.5 & 58.1 & 69.2 & 83.6 \\
        BLOOM & 53.0 & 48.9 & 59.3 & 72.9 & 84.7 \\
        BLOOM-7B1 & 59.5 & 48.5 & 57.5 & 67.5 & 85.2 \\
        BLOOM-3B & 59.5 & 49.3 & 59.9 & 65.7 & 85.3 \\
        BLOOM-1B7 & 63.3 & 46.2 & 56.6 & 63.9 & 83.4 \\
        BLOOM-1B1 & 60.8 & 44.7 & 54.9 & 58.6 & 82.3 \\
        OPT-175B & 50.3 & 50.5 & 60.0 & 65.2 & 86.1 \\
        OPT-66B & 53.5 & 50.9 & 57.5 & 65.1 & 84.5 \\
        OPT-30B & 58.1 & 49.8 & 57.6 & 66.6 & 85.4 \\
        OPT-13B & 54.6 & 51.3 & 56.6 & 65.3 & 83.7 \\
        OPT-6.7B & 56.3 & 50.5 & 55.5 & 65.3 & 84.3 \\
        OPT-2.7B & 56.6 & 52.1 & 55.4 & 66.2 & 84.2 \\
        OPT-1.3B & 57.2 & 48.9 & 54.0 & 64.7 & 82.6 \\
        \bottomrule
    \end{tabular}
    \caption{The performance of the models on XSum with various alternative-choices using PMI as the scoring function.}
    \label{tab:xsum_total_pmi}
\end{table*}

\begin{table*}[!ht]
    \small
    \centering
    \begin{tabular}{@{} l c c c c c @{}}
        \toprule
        Model & FIR & FCMG & FIB & FactCC & MFMA \\
       \midrule
        T0-3B & 45.2 & 15.9 & 34.4 & 98.5 & 73.5 \\
        T0 & 34.7 & 23.0 & 38.9 & 97.9 & 78.0 \\
        FLAN-T5-xl & 52.8 & 18.5 & 35.6 & 98.3 & 74.9 \\
        FLAN-T5-xxl & 49.4 & 18.5 & 39.2 & 98.3 & 78.1 \\
        T5-LM-Adapt-xl & 82.6 & 23.8 & 44.6 & 98.1 & 71.4 \\
        T5-LM-Adapt-xxl & 72.2 & 22.0 & 43.4 & 98.3 & 75.1 \\
        GPT-Neo-1.3B & 83.3 & 22.2 & 46.9 & 97.0 & 66.1 \\
        GPT2-XL & 78.6 & 22.1 & 45.6 & 97.3 & 67.9 \\
        GPT-Neo-2.7B & 81.3 & 23.1 & 46.8 & 97.1 & 67.6 \\
        GPTJ-6B & 72.2 & 22.9 & 47.2 & 98.0 & 74.6 \\
        GPT-Neox-20B & 68.2 & 26.9 & 47.7 & 97.9 & 75.9 \\
        BLOOM & 70.6 & 24.5 & 48.6 & 98.5 & 78.8 \\
        BLOOM-7B1 & 81.7 & 24.4 & 48.4 & 97.6 & 71.9 \\
        BLOOM-3B & 85.1 & 24.4 & 48.6 & 97.3 & 68.5 \\
        BLOOM-1B7 & 87.3 & 25.4 & 48.5 & 96.2 & 65.1 \\
        BLOOM-1B1 & 90.4 & 24.7 & 49.3 & 96.2 & 64.2 \\
        OPT-175B & 53.2 & 26.4 & 48.1 & 98.3 & 81.8 \\
        OPT-66B & 61.0 & 25.5 & 47.4 & 98.3 & 80.2 \\
        OPT-30B & 60.6 & 25.6 & 47.0 & 98.1 & 78.3 \\
        OPT-13B & 66.8 & 24.6 & 46.3 & 98.1 & 78.8 \\
        OPT-6.7B & 66.1 & 25.9 & 45.6 & 97.6 & 75.7 \\
        OPT-2.7B & 72.6 & 24.6 & 45.7 & 98.1 & 73.2 \\
        OPT-1.3B & 77.3 & 23.1 & 45.2 & 97.4 & 71.8 \\
        \bottomrule
    \end{tabular}
    \caption{The performance of the models on XSum with various alternative-choices using LL as the scoring function.}
    \label{tab:xsum_total_ll}
\end{table*}

\begin{table*}[!ht]
    \small
    \centering
    \begin{tabular}{@{} l c c c c c @{}}
        \toprule
        Model & FIR & FCMG & FIB & FactCC & MFMA \\
       \midrule
        T0-3B & 65.6 & 7.0 & 17.7 & 82.4 & 98.0 \\
        T0 & 50.0 & 4.4 & 11.4 & 79.9 & 92.0 \\
        FLAN-T5-xl & 65.6 & 7.4 & 16.0 & 79.7 & 100.0 \\
        FLAN-T5-xxl & 59.4 & 6.3 & 13.8 & 76.5 & 100.0 \\
        T5-LM-Adapt-xl & 62.5 & 4.9 & 12.7 & 79.6 & 99.0 \\
        T5-LM-Adapt-xxl & 59.4 & 6.0 & 12.0 & 76.8 & 99.0 \\
        GPT-Neo-1.3B & 78.1 & 6.4 & 8.7 & 77.7 & 100.0 \\
        GPT2-XL & 78.1 & 8.2 & 9.8 & 79.5 & 99.0 \\
        GPT-Neo-2.7B & 78.1 & 7.9 & 10.1 & 78.2 & 99.0 \\
        GPTJ-6B & 78.1 & 7.5 & 8.1 & 82.0 & 99.0 \\
        GPT-Neox-20B & 71.9 & 8.6 & 10.5 & 76.2 & 97.0 \\
        BLOOM & 75.0 & 10.8 & 9.2 & 79.3 & 99.0 \\
        BLOOM-7B1 & 84.4 & 9.8 & 10.3 & 81.8 & 99.0 \\
        BLOOM-3B & 78.1 & 8.0 & 7.9 & 78.2 & 100.0 \\
        BLOOM-1B7 & 84.4 & 6.8 & 9.2 & 76.3 & 99.0 \\
        BLOOM-1B1 & 84.4 & 7.5 & 11.2 & 75.8 & 100.0 \\
        OPT-175B & 71.9 & 11.9 & 10.7 & 75.2 & 98.0 \\
        OPT-66B & 71.9 & 8.8 & 9.2 & 75.9 & 99.0 \\
        OPT-30B & 71.9 & 11.1 & 9.0 & 77.3 & 100.0 \\
        OPT-13B & 75.0 & 8.2 & 9.6 & 79.5 & 99.0 \\
        OPT-6.7B & 81.2 & 10.2 & 9.9 & 79.8 & 99.0 \\
        OPT-2.7B & 75.0 & 7.8 & 9.6 & 74.1 & 98.0 \\
        OPT-1.3B & 78.1 & 6.8 & 8.1 & 75.3 & 100.0 \\
        \bottomrule
    \end{tabular}
    \caption{The performance of the models on CNN/DM with various alternative-choices using avg. PMI as the scoring function.}
    \label{tab:cnn_dm_total_avg_pmi}
\end{table*}

\begin{table*}[!ht]
    \small
    \centering
    \begin{tabular}{@{} l c c c c c @{}}
        \toprule
        Model & FIR & FCMG & FIB & FactCC & MFMA \\
       \midrule
        T0-3B & 40.6 & 3.3 & 11.6 & 90.3 & 100.0 \\
        T0 & 37.5 & 2.2 & 8.3 & 90.8 & 100.0 \\
        FLAN-T5-xl & 40.6 & 1.7 & 9.0 & 91.4 & 100.0 \\
        FLAN-T5-xxl & 40.6 & 1.1 & 6.1 & 88.9 & 100.0 \\
        T5-LM-Adapt-xl & 40.6 & 1.6 & 6.6 & 88.2 & 99.0 \\
        T5-LM-Adapt-xxl & 31.2 & 1.2 & 5.3 & 89.8 & 100.0 \\
        GPT-Neo-1.3B & 46.9 & 0.7 & 1.3 & 93.6 & 99.0 \\
        GPT2-XL & 56.2 & 0.9 & 2.6 & 92.5 & 99.0 \\
        GPT-Neo-2.7B & 50.0 & 0.8 & 1.8 & 92.9 & 97.0 \\
        GPTJ-6B & 46.9 & 0.5 & 2.0 & 95.2 & 99.0 \\
        GPT-Neox-20B & 40.6 & 0.2 & 1.8 & 94.2 & 98.0 \\
        BLOOM & 40.6 & 0.3 & 1.8 & 93.8 & 99.0 \\
        BLOOM-7B1 & 50.0 & 1.0 & 2.8 & 95.9 & 100.0 \\
        BLOOM-3B & 53.1 & 1.2 & 2.2 & 93.5 & 100.0 \\
        BLOOM-1B7 & 53.1 & 0.9 & 2.2 & 92.9 & 99.0 \\
        BLOOM-1B1 & 62.5 & 1.3 & 2.6 & 93.6 & 98.0 \\
        OPT-175B & 40.6 & 0.6 & 2.2 & 91.4 & 99.0 \\
        OPT-66B & 43.8 & 0.9 & 2.2 & 92.8 & 99.0 \\
        OPT-30B & 43.8 & 0.8 & 2.0 & 94.1 & 99.0 \\
        OPT-13B & 43.8 & 0.9 & 1.8 & 95.5 & 99.0 \\
        OPT-6.7B & 56.2 & 0.9 & 2.6 & 94.6 & 98.0 \\
        OPT-2.7B & 43.8 & 1.2 & 2.6 & 92.9 & 98.0 \\
        OPT-1.3B & 46.9 & 1.2 & 2.0 & 92.5 & 98.0 \\
        \bottomrule
    \end{tabular}
    \caption{The performance of the models on CNN/DM with various alternative-choices using avg. LL as the scoring function.}
    \label{tab:cnn_dm_total_avg_ll}
\end{table*}

\begin{table*}[!ht]
    \small
    \centering
    \begin{tabular}{@{} l c c c c c @{}}
        \toprule
        Model & FIR & FCMG & FIB & FactCC & MFMA \\
       \midrule
        T0-3B & 46.9 & 1.6 & 8.5 & 76.6 & 100.0 \\
        T0 & 28.1 & 1.2 & 6.1 & 75.9 & 96.0 \\
        FLAN-T5-xl & 40.6 & 1.6 & 7.2 & 74.6 & 100.0 \\
        FLAN-T5-xxl & 34.4 & 1.7 & 5.9 & 69.9 & 100.0 \\
        T5-LM-Adapt-xl & 34.4 & 1.1 & 6.1 & 69.4 & 98.0 \\
        T5-LM-Adapt-xxl & 34.4 & 0.9 & 5.3 & 68.4 & 99.0 \\
        GPT-Neo-1.3B & 50.0 & 0.5 & 3.7 & 69.8 & 99.0 \\
        GPT2-XL & 43.8 & 0.4 & 3.5 & 69.8 & 99.0 \\
        GPT-Neo-2.7B & 46.9 & 0.4 & 2.6 & 66.9 & 99.0 \\
        GPTJ-6B & 59.4 & 0.5 & 2.4 & 73.6 & 99.0 \\
        GPT-Neox-20B & 56.2 & 0.4 & 2.4 & 69.0 & 99.0 \\
        BLOOM & 40.6 & 0.5 & 2.4 & 69.7 & 99.0 \\
        BLOOM-7B1 & 56.2 & 0.5 & 2.9 & 73.9 & 100.0 \\
        BLOOM-3B & 56.2 & 0.5 & 2.9 & 71.1 & 100.0 \\
        BLOOM-1B7 & 53.1 & 0.5 & 3.3 & 64.8 & 98.0 \\
        BLOOM-1B1 & 59.4 & 0.5 & 3.5 & 68.4 & 99.0 \\
        OPT-175B & 53.1 & 0.7 & 2.8 & 70.4 & 98.0 \\
        OPT-66B & 59.4 & 0.5 & 2.4 & 68.1 & 99.0 \\
        OPT-30B & 53.1 & 0.6 & 3.1 & 71.9 & 99.0 \\
        OPT-13B & 43.8 & 0.6 & 3.1 & 71.3 & 98.0 \\
        OPT-6.7B & 53.1 & 0.5 & 2.4 & 72.6 & 99.0 \\
        OPT-2.7B & 56.2 & 0.5 & 3.1 & 66.0 & 98.0 \\
        OPT-1.3B & 53.1 & 0.5 & 3.7 & 69.3 & 99.0 \\
        \bottomrule
    \end{tabular}
    \caption{The performance of the models on CNN/DM with various alternative-choices using PMI as the scoring function.}
    \label{tab:cnn_dm_total_pmi}
\end{table*}

\begin{table*}[!ht]
    \small
    \centering
    \begin{tabular}{@{} l c c c c c @{}}
        \toprule
        Model & FIR & FCMG & FIB & FactCC & MFMA \\
       \midrule
        T0-3B & 71.9 & 45.1 & 52.7 & 98.7 & 97.0 \\
        T0 & 62.5 & 37.4 & 42.7 & 97.4 & 97.0 \\
        FLAN-T5-xl & 75.0 & 42.8 & 48.6 & 98.4 & 98.0 \\
        FLAN-T5-xxl & 68.8 & 26.9 & 35.5 & 97.0 & 99.0 \\
        T5-LM-Adapt-xl & 90.6 & 39.7 & 45.1 & 97.0 & 89.0 \\
        T5-LM-Adapt-xxl & 68.8 & 31.4 & 32.6 & 98.7 & 94.0 \\
        GPT-Neo-1.3B & 78.1 & 24.3 & 20.1 & 97.4 & 99.0 \\
        GPT2-XL & 81.2 & 26.9 & 26.5 & 96.6 & 97.0 \\
        GPT-Neo-2.7B & 75.0 & 24.1 & 19.9 & 97.0 & 98.0 \\
        GPTJ-6B & 78.1 & 21.0 & 18.6 & 97.9 & 99.0 \\
        GPT-Neox-20B & 75.0 & 22.5 & 20.4 & 98.0 & 99.0 \\
        BLOOM & 59.4 & 16.7 & 16.6 & 98.3 & 100.0 \\
        BLOOM-7B1 & 78.1 & 22.1 & 21.0 & 97.6 & 100.0 \\
        BLOOM-3B & 78.1 & 25.2 & 20.6 & 98.0 & 98.0 \\
        BLOOM-1B7 & 81.2 & 23.4 & 20.1 & 97.0 & 98.0 \\
        BLOOM-1B1 & 84.4 & 26.2 & 23.2 & 97.4 & 98.0 \\
        OPT-175B & 65.6 & 25.9 & 20.8 & 97.3 & 99.0 \\
        OPT-66B & 68.8 & 26.7 & 23.6 & 97.9 & 99.0 \\
        OPT-30B & 75.0 & 25.3 & 21.0 & 97.9 & 100.0 \\
        OPT-13B & 68.8 & 28.1 & 24.3 & 97.9 & 100.0 \\
        OPT-6.7B & 78.1 & 29.4 & 26.7 & 98.7 & 100.0 \\
        OPT-2.7B & 71.9 & 29.5 & 25.8 & 98.3 & 100.0 \\
        OPT-1.3B & 75.0 & 27.8 & 23.8 & 98.3 & 100.0 \\
        \bottomrule
    \end{tabular}
    \caption{The performance of the models on CNN/DM with various alternative-choices using LL as the scoring function.}
    \label{tab:cnn_dm_total_ll}
\end{table*}

\section{Accuracies from MFMA-Generated Summaries}
\label{sec:mfma_scores}
We show the performance of different models on MFMA-generated summaries broken down by the model used to generate the summary for XSum using different scoring functions in \cref{tab:xsum_mfma_avg_pmi}, \cref{tab:xsum_mfma_avg_ll}, \cref{tab:xsum_mfma_pmi}, and \cref{tab:xsum_mfma_ll}.

\begin{table*}[!ht]
    \small
    \centering
    \begin{tabular}{@{} l c c @{}}
        \toprule
        Model & BART-base & T5-base \\
       \midrule
        T0-3B & 93.4 & 74.9 \\
        T0 & 94.2 & 71.2 \\
        FLAN-T5-xl & 94.8 & 74.3 \\
        FLAN-T5-xxl & 95.0 & 77.9 \\
        T5-LM-Adapt-xl & 94.2 & 79.3 \\
        T5-LM-Adapt-xxl & 95.0 & 81.0 \\
        GPT-Neo-1.3B & 93.6 & 79.1 \\
        GPT2-XL & 91.7 & 72.9 \\
        GPT-Neo-2.7B & 94.4 & 73.7 \\
        GPTJ-6B & 94.2 & 78.8 \\
        GPT-Neox-20B & 95.2 & 75.7 \\
        BLOOM & 95.0 & 79.6 \\
        BLOOM-7B1 & 94.6 & 76.5 \\
        BLOOM-3B & 94.4 & 77.1 \\
        BLOOM-1B7 & 95.0 & 77.4 \\
        BLOOM-1B1 & 93.2 & 75.7 \\
        OPT-175B & 94.6 & 77.7 \\
        OPT-66B & 95.2 & 77.4 \\
        OPT-30B & 94.8 & 77.9 \\
        OPT-13B & 95.0 & 76.0 \\
        OPT-6.7B & 95.0 & 75.7 \\
        OPT-2.7B & 94.0 & 75.7 \\
        OPT-1.3B & 93.8 & 76.5 \\
        \bottomrule
    \end{tabular}
    \caption{The performance of the models on XSum with MFMA-generated alternative-choices using avg. PMI as the scoring function.}
    \label{tab:xsum_mfma_avg_pmi}
\end{table*}

\begin{table*}[!ht]
    \small
    \centering
    \begin{tabular}{@{} l c c @{}}
        \toprule
        Model & BART-base & T5-base \\
       \midrule
        T0-3B & 79.7 & 54.2 \\
        T0 & 83.0 & 61.2 \\
        FLAN-T5-xl & 81.0 & 54.2 \\
        FLAN-T5-xxl & 82.8 & 58.7 \\
        T5-LM-Adapt-xl & 71.2 & 51.4 \\
        T5-LM-Adapt-xxl & 74.9 & 57.3 \\
        GPT-Neo-1.3B & 65.6 & 51.1 \\
        GPT2-XL & 66.5 & 53.6 \\
        GPT-Neo-2.7B & 69.6 & 52.8 \\
        GPTJ-6B & 76.8 & 59.2 \\
        GPT-Neox-20B & 76.0 & 62.3 \\
        BLOOM & 80.1 & 64.5 \\
        BLOOM-7B1 & 72.3 & 57.5 \\
        BLOOM-3B & 71.4 & 54.7 \\
        BLOOM-1B7 & 69.4 & 51.1 \\
        BLOOM-1B1 & 67.9 & 49.7 \\
        OPT-175B & 83.0 & 69.9 \\
        OPT-66B & 81.8 & 67.0 \\
        OPT-30B & 78.7 & 64.8 \\
        OPT-13B & 79.5 & 65.6 \\
        OPT-6.7B & 76.0 & 63.1 \\
        OPT-2.7B & 74.1 & 62.0 \\
        OPT-1.3B & 70.8 & 57.3 \\
        \bottomrule
    \end{tabular}
    \caption{The performance of the models on XSum with MFMA-generated alternative-choices using avg. LL as the scoring function.}
    \label{tab:xsum_mfma_avg_ll}
\end{table*}

\begin{table*}[!ht]
    \small
    \centering
    \begin{tabular}{@{} l c c @{}}
        \toprule
        Model & BART-base & T5-base \\
       \midrule
        T0-3B & 93.6 & 71.5 \\
        T0 & 94.2 & 70.9 \\
        FLAN-T5-xl & 93.2 & 70.4 \\
        FLAN-T5-xxl & 94.4 & 74.9 \\
        T5-LM-Adapt-xl & 91.9 & 74.0 \\
        T5-LM-Adapt-xxl & 93.6 & 74.6 \\
        GPT-Neo-1.3B & 92.3 & 73.7 \\
        GPT2-XL & 91.1 & 66.8 \\
        GPT-Neo-2.7B & 92.3 & 70.1 \\
        GPTJ-6B & 93.2 & 73.5 \\
        GPT-Neox-20B & 93.4 & 71.5 \\
        BLOOM & 93.2 & 74.3 \\
        BLOOM-7B1 & 93.8 & 74.6 \\
        BLOOM-3B & 94.0 & 74.6 \\
        BLOOM-1B7 & 93.4 & 71.2 \\
        BLOOM-1B1 & 91.7 & 70.7 \\
        OPT-175B & 94.0 & 76.5 \\
        OPT-66B & 93.4 & 73.7 \\
        OPT-30B & 94.4 & 74.3 \\
        OPT-13B & 94.2 & 70.9 \\
        OPT-6.7B & 93.0 & 73.7 \\
        OPT-2.7B & 93.6 & 72.6 \\
        OPT-1.3B & 92.1 & 70.9 \\
        \bottomrule
    \end{tabular}
    \caption{The performance of the models on MFMA-generated alternative-choices using PMI as the scoring function.}
    \label{tab:xsum_mfma_pmi}
\end{table*}

\begin{table*}[!ht]
    \small
    \centering
    \begin{tabular}{@{} l c c @{}}
        \toprule
        Model & BART-base & T5-base \\
       \midrule
        T0-3B & 85.9 & 58.4 \\
        T0 & 88.2 & 65.6 \\
        FLAN-T5-xl & 87.4 & 59.5 \\
        FLAN-T5-xxl & 89.6 & 64.0 \\
        T5-LM-Adapt-xl & 80.3 & 60.6 \\
        T5-LM-Adapt-xxl & 84.7 & 63.4 \\
        GPT-Neo-1.3B & 73.3 & 57.3 \\
        GPT2-XL & 75.4 & 58.7 \\
        GPT-Neo-2.7B & 75.8 & 57.5 \\
        GPTJ-6B & 83.2 & 64.0 \\
        GPT-Neox-20B & 83.2 & 67.0 \\
        BLOOM & 86.3 & 69.6 \\
        BLOOM-7B1 & 78.3 & 64.0 \\
        BLOOM-3B & 76.4 & 58.9 \\
        BLOOM-1B7 & 72.0 & 56.7 \\
        BLOOM-1B1 & 72.3 & 54.2 \\
        OPT-175B & 88.6 & 73.5 \\
        OPT-66B & 86.1 & 72.9 \\
        OPT-30B & 86.1 & 68.7 \\
        OPT-13B & 86.1 & 69.8 \\
        OPT-6.7B & 84.3 & 65.1 \\
        OPT-2.7B & 81.2 & 63.4 \\
        OPT-1.3B & 78.5 & 63.7 \\
        \bottomrule
    \end{tabular}
    \caption{The performance of the models on XSum with MFMA-generated alternative-choices using LL as the scoring function.}
    \label{tab:xsum_mfma_ll}
\end{table*}

\section{Accuracies from FactCC-Generated Summaries}
\label{sec:factcc_scores}
We show the performance of different models on FactCC-generated summaries broken down by the method used to generate the summary using different scoring functions for XSum in \cref{tab:xsum_factcc_avg_pmi}, \cref{tab:xsum_factcc_avg_ll}, \cref{tab:xsum_factcc_pmi}, \cref{tab:xsum_factcc_ll} and for CNN/DM in  \cref{tab:cnn_dm_factcc_avg_pmi}, \cref{tab:cnn_dm_factcc_avg_ll}, \cref{tab:cnn_dm_factcc_pmi}, \cref{tab:cnn_dm_factcc_ll}. 
\begin{table*}[!ht]
    \small
    \centering
    \begin{tabular}{@{} l c c c c c @{}}
        \toprule
        Model & Date Swap & Entity Swap & Negation & Number Swap & Pronoun \\
       \midrule
        T0-3B & 76.4 & 86.6 & 94.5 & 76.5 & 78.7 \\
        T0 & 85.5 & 86.9 & 93.9 & 92.6 & 84.8 \\
        FLAN-T5-xl & 72.7 & 86.0 & 96.1 & 82.4 & 72.6 \\
        FLAN-T5-xxl & 76.4 & 85.5 & 97.2 & 85.3 & 67.1 \\
        T5-LM-Adapt-xl & 67.3 & 75.9 & 89.9 & 60.3 & 65.2 \\
        T5-LM-Adapt-xxl & 69.1 & 81.4 & 94.5 & 70.6 & 72.0 \\
        GPT-Neo-1.3B & 52.7 & 66.3 & 75.5 & 42.6 & 72.0 \\
        GPT2-XL & 60.0 & 69.2 & 82.1 & 41.2 & 63.4 \\
        GPT-Neo-2.7B & 65.5 & 65.7 & 81.2 & 54.4 & 70.7 \\
        GPTJ-6B & 60.0 & 70.6 & 85.1 & 54.4 & 63.4 \\
        GPT-Neox-20B & 61.8 & 68.9 & 86.2 & 55.9 & 62.8 \\
        BLOOM & 60.0 & 72.1 & 83.4 & 67.6 & 66.5 \\
        BLOOM-7B1 & 60.0 & 71.5 & 76.8 & 52.9 & 65.9 \\
        BLOOM-3B & 50.9 & 69.5 & 75.7 & 57.4 & 69.5 \\
        BLOOM-1B7 & 54.5 & 65.1 & 70.5 & 60.3 & 73.8 \\
        BLOOM-1B1 & 58.2 & 63.1 & 65.9 & 54.4 & 66.5 \\
        OPT-175B & 56.4 & 64.8 & 83.2 & 61.8 & 59.8 \\
        OPT-66B & 58.2 & 63.7 & 84.0 & 60.3 & 57.3 \\
        OPT-30B & 61.8 & 65.1 & 84.5 & 63.2 & 59.1 \\
        OPT-13B & 65.5 & 68.6 & 81.6 & 63.2 & 55.5 \\
        OPT-6.7B & 63.6 & 66.9 & 80.1 & 60.3 & 57.9 \\
        OPT-2.7B & 60.0 & 65.1 & 82.7 & 51.5 & 59.1 \\
        OPT-1.3B & 63.6 & 63.1 & 83.2 & 57.4 & 58.5 \\
        \bottomrule
    \end{tabular}
    \caption{The performance of the models on XSum with FactCC-generated alternative-choices using avg. PMI as the scoring function.}
    \label{tab:xsum_factcc_avg_pmi}
\end{table*}

\begin{table*}[!ht]
    \small
    \centering
    \begin{tabular}{@{} l c c c c c @{}}
        \toprule
        Model & Date Swap & Entity Swap & Negation & Number Swap & Pronoun \\
       \midrule
        T0-3B & 96.4 & 96.5 & 98.7 & 94.1 & 99.4 \\
        T0 & 100.0 & 95.3 & 96.7 & 97.1 & 99.4 \\
        FLAN-T5-xl & 100.0 & 96.2 & 98.7 & 92.6 & 99.4 \\
        FLAN-T5-xxl & 98.2 & 95.9 & 99.1 & 98.5 & 99.4 \\
        T5-LM-Adapt-xl & 92.7 & 91.0 & 92.8 & 89.7 & 100.0 \\
        T5-LM-Adapt-xxl & 94.5 & 93.3 & 96.9 & 89.7 & 100.0 \\
        GPT-Neo-1.3B & 96.4 & 89.5 & 97.6 & 88.2 & 99.4 \\
        GPT2-XL & 96.4 & 91.3 & 97.8 & 86.8 & 100.0 \\
        GPT-Neo-2.7B & 96.4 & 92.4 & 98.2 & 86.8 & 100.0 \\
        GPTJ-6B & 98.2 & 93.9 & 98.9 & 88.2 & 100.0 \\
        GPT-Neox-20B & 98.2 & 93.6 & 99.3 & 89.7 & 100.0 \\
        BLOOM & 98.2 & 95.3 & 99.6 & 92.6 & 100.0 \\
        BLOOM-7B1 & 98.2 & 92.7 & 99.1 & 85.3 & 100.0 \\
        BLOOM-3B & 92.7 & 91.6 & 99.1 & 85.3 & 100.0 \\
        BLOOM-1B7 & 92.7 & 89.8 & 98.5 & 83.8 & 99.4 \\
        BLOOM-1B1 & 90.9 & 86.9 & 96.7 & 85.3 & 99.4 \\
        OPT-175B & 100.0 & 95.6 & 99.3 & 92.6 & 100.0 \\
        OPT-66B & 98.2 & 94.8 & 99.6 & 89.7 & 100.0 \\
        OPT-30B & 98.2 & 95.1 & 98.9 & 91.2 & 100.0 \\
        OPT-13B & 98.2 & 94.8 & 97.8 & 88.2 & 100.0 \\
        OPT-6.7B & 98.2 & 95.1 & 98.5 & 83.8 & 100.0 \\
        OPT-2.7B & 98.2 & 93.9 & 98.9 & 86.8 & 100.0 \\
        OPT-1.3B & 96.4 & 91.9 & 98.5 & 89.7 & 99.4 \\
        \bottomrule
    \end{tabular}
    \caption{The performance of the models on XSum with FactCC-generated alternative-choices using avg. LL as the scoring function.}
    \label{tab:xsum_factcc_avg_ll}
\end{table*}

\begin{table*}[!ht]
    \small
    \centering
    \begin{tabular}{@{} l c c c c c @{}}
        \toprule
        Model & Date Swap & Entity Swap & Negation & Number Swap & Pronoun \\
       \midrule
        T0-3B & 83.6 & 83.7 & 84.2 & 80.9 & 80.5 \\
        T0 & 87.3 & 86.0 & 92.3 & 91.2 & 86.0 \\
        FLAN-T5-xl & 80.0 & 78.8 & 87.1 & 83.8 & 74.4 \\
        FLAN-T5-xxl & 78.2 & 79.9 & 86.2 & 86.8 & 69.5 \\
        T5-LM-Adapt-xl & 70.9 & 70.9 & 69.8 & 64.7 & 70.1 \\
        T5-LM-Adapt-xxl & 74.5 & 75.0 & 79.9 & 72.1 & 75.0 \\
        GPT-Neo-1.3B & 63.6 & 63.4 & 57.1 & 38.2 & 72.0 \\
        GPT2-XL & 65.5 & 64.0 & 68.5 & 42.6 & 63.4 \\
        GPT-Neo-2.7B & 65.5 & 64.8 & 67.8 & 54.4 & 70.7 \\
        GPTJ-6B & 69.1 & 66.9 & 69.4 & 52.9 & 63.4 \\
        GPT-Neox-20B & 65.5 & 66.0 & 76.4 & 55.9 & 62.8 \\
        BLOOM & 65.5 & 69.5 & 79.9 & 64.7 & 66.5 \\
        BLOOM-7B1 & 63.6 & 67.4 & 71.3 & 50.0 & 65.9 \\
        BLOOM-3B & 58.2 & 65.4 & 67.4 & 52.9 & 69.5 \\
        BLOOM-1B7 & 54.5 & 63.7 & 63.2 & 52.9 & 73.8 \\
        BLOOM-1B1 & 58.2 & 59.9 & 56.2 & 50.0 & 66.5 \\
        OPT-175B & 54.5 & 61.9 & 71.1 & 64.7 & 59.8 \\
        OPT-66B & 67.3 & 58.7 & 73.3 & 60.3 & 57.3 \\
        OPT-30B & 61.8 & 62.5 & 73.3 & 64.7 & 59.1 \\
        OPT-13B & 67.3 & 64.5 & 69.4 & 63.2 & 55.5 \\
        OPT-6.7B & 67.3 & 62.8 & 70.7 & 57.4 & 57.9 \\
        OPT-2.7B & 63.6 & 65.4 & 72.2 & 50.0 & 59.1 \\
        OPT-1.3B & 67.3 & 60.5 & 71.1 & 55.9 & 58.5 \\
        \bottomrule
    \end{tabular}
    \caption{The performance of the models on XSum with FactCC-generated alternative-choices using PMI as the scoring function.}
    \label{tab:xsum_factcc_pmi}
\end{table*}

\begin{table*}[!ht]
    \small
    \centering
    \begin{tabular}{@{} l c c c c c @{}}
        \toprule
        Model & Date Swap & Entity Swap & Negation & Number Swap & Pronoun \\
       \midrule
        T0-3B & 98.2 & 96.8 & 100.0 & 95.6 & 99.4 \\
        T0 & 98.2 & 95.6 & 99.1 & 98.5 & 98.8 \\
        FLAN-T5-xl & 100.0 & 96.2 & 100.0 & 94.1 & 99.4 \\
        FLAN-T5-xxl & 98.2 & 95.6 & 100.0 & 98.5 & 99.4 \\
        T5-LM-Adapt-xl & 98.2 & 95.9 & 100.0 & 91.2 & 100.0 \\
        T5-LM-Adapt-xxl & 98.2 & 96.8 & 100.0 & 89.7 & 100.0 \\
        GPT-Neo-1.3B & 96.4 & 93.9 & 99.8 & 88.2 & 99.4 \\
        GPT2-XL & 96.4 & 95.1 & 99.6 & 86.8 & 100.0 \\
        GPT-Neo-2.7B & 96.4 & 94.8 & 99.1 & 88.2 & 100.0 \\
        GPTJ-6B & 98.2 & 96.2 & 100.0 & 88.2 & 100.0 \\
        GPT-Neox-20B & 98.2 & 95.9 & 99.8 & 89.7 & 100.0 \\
        BLOOM & 100.0 & 97.1 & 99.8 & 91.2 & 100.0 \\
        BLOOM-7B1 & 98.2 & 95.3 & 100.0 & 86.8 & 100.0 \\
        BLOOM-3B & 92.7 & 94.8 & 100.0 & 88.2 & 100.0 \\
        BLOOM-1B7 & 90.9 & 93.0 & 99.3 & 88.2 & 99.4 \\
        BLOOM-1B1 & 94.5 & 92.2 & 99.6 & 86.8 & 99.4 \\
        OPT-175B & 100.0 & 96.2 & 99.8 & 92.6 & 100.0 \\
        OPT-66B & 98.2 & 97.1 & 100.0 & 89.7 & 100.0 \\
        OPT-30B & 98.2 & 96.5 & 99.6 & 91.2 & 100.0 \\
        OPT-13B & 100.0 & 96.8 & 99.8 & 86.8 & 100.0 \\
        OPT-6.7B & 100.0 & 96.2 & 99.6 & 83.8 & 100.0 \\
        OPT-2.7B & 100.0 & 96.5 & 100.0 & 86.8 & 100.0 \\
        OPT-1.3B & 98.2 & 94.8 & 100.0 & 88.2 & 99.4 \\
        \bottomrule
    \end{tabular}
    \caption{The performance of the models on XSum with FactCC-generated alternative-choices using LL as the scoring function.}
    \label{tab:xsum_factcc_ll}
\end{table*}

\begin{table*}[!ht]
    \small
    \centering
    \begin{tabular}{@{} l c c c c c @{}}
        \toprule
        Model & Date Swap & Entity Swap & Negation & Number Swap & Pronoun \\
       \midrule
        T0-3B & 81.8 & 78.3 & 91.6 & 75.0 & 80.0 \\
        T0 & 81.8 & 73.9 & 94.0 & 66.7 & 73.3 \\
        flan-t5-xl & 78.2 & 75.4 & 92.8 & 77.8 & 66.7 \\
        flan-t5-xxl & 76.4 & 71.0 & 90.4 & 69.4 & 66.7 \\
        t5-lm-adapt-xl & 80.0 & 81.2 & 84.3 & 75.0 & 71.1 \\
        t5-lm-adapt-xxl & 80.0 & 71.0 & 86.7 & 75.0 & 66.7 \\
        GPT-Neo-1.3B & 72.7 & 75.4 & 85.5 & 75.0 & 75.6 \\
        GPT2-XL & 78.2 & 79.7 & 86.7 & 75.0 & 71.1 \\
        GPT-Neo-2.7B & 74.5 & 73.9 & 85.5 & 80.6 & 75.6 \\
        GPTJ-6B & 80.0 & 76.8 & 91.6 & 83.3 & 75.6 \\
        GPT-Neox-20B & 67.3 & 72.5 & 88.0 & 77.8 & 71.1 \\
        BLOOM & 80.0 & 75.4 & 85.5 & 77.8 & 75.6 \\
        BLOOM-7B1 & 81.8 & 78.3 & 84.3 & 80.6 & 84.4 \\
        BLOOM-3B & 80.0 & 79.7 & 75.9 & 80.6 & 75.6 \\
        BLOOM-1B7 & 78.2 & 73.9 & 77.1 & 77.8 & 75.6 \\
        BLOOM-1B1 & 80.0 & 71.0 & 78.3 & 77.8 & 73.3 \\
        OPT-175B & 70.9 & 72.5 & 84.3 & 75.0 & 68.9 \\
        OPT-66B & 69.1 & 72.5 & 83.1 & 75.0 & 77.8 \\
        OPT-30B & 74.5 & 68.1 & 88.0 & 77.8 & 77.8 \\
        OPT-13B & 80.0 & 78.3 & 84.3 & 72.2 & 77.8 \\
        OPT-6.7B & 76.4 & 84.1 & 88.0 & 66.7 & 71.1 \\
        OPT-2.7B & 65.5 & 76.8 & 81.9 & 69.4 & 68.9 \\
        OPT-1.3B & 72.7 & 75.4 & 79.5 & 72.2 & 73.3 \\
        \bottomrule
    \end{tabular}
    \caption{The performance of the models on CNN/DM with FactCC-generated alternative-choices using avg. PMI as the scoring function.}
    \label{tab:cnn_dm_factcc_avg_pmi}
\end{table*}

\begin{table*}[!ht]
    \small
    \centering
    \begin{tabular}{@{} l c c c c c @{}}
        \toprule
        Model & Date Swap & Entity Swap & Negation & Number Swap & Pronoun \\
       \midrule
        T0-3B & 92.7 & 89.9 & 91.6 & 86.1 & 88.9 \\
        T0 & 92.7 & 92.8 & 94.0 & 80.6 & 86.7 \\
        flan-t5-xl & 94.5 & 92.8 & 91.6 & 86.1 & 88.9 \\
        flan-t5-xxl & 92.7 & 88.4 & 94.0 & 80.6 & 82.2 \\
        t5-lm-adapt-xl & 89.1 & 88.4 & 89.2 & 86.1 & 86.7 \\
        t5-lm-adapt-xxl & 90.9 & 92.8 & 88.0 & 88.9 & 86.7 \\
        GPT-Neo-1.3B & 87.3 & 97.1 & 97.6 & 86.1 & 93.3 \\
        GPT2-XL & 87.3 & 94.2 & 95.2 & 88.9 & 93.3 \\
        GPT-Neo-2.7B & 89.1 & 95.7 & 94.0 & 91.7 & 91.1 \\
        GPTJ-6B & 92.7 & 95.7 & 97.6 & 91.7 & 95.6 \\
        GPT-Neox-20B & 90.9 & 95.7 & 96.4 & 91.7 & 93.3 \\
        BLOOM & 92.7 & 94.2 & 95.2 & 88.9 & 95.6 \\
        BLOOM-7B1 & 92.7 & 97.1 & 98.8 & 91.7 & 95.6 \\
        BLOOM-3B & 94.5 & 95.7 & 95.2 & 83.3 & 93.3 \\
        BLOOM-1B7 & 92.7 & 95.7 & 94.0 & 86.1 & 91.1 \\
        BLOOM-1B1 & 90.9 & 97.1 & 95.2 & 86.1 & 93.3 \\
        OPT-175B & 89.1 & 92.8 & 94.0 & 91.7 & 86.7 \\
        OPT-66B & 87.3 & 94.2 & 95.2 & 91.7 & 93.3 \\
        OPT-30B & 89.1 & 94.2 & 97.6 & 94.4 & 93.3 \\
        OPT-13B & 94.5 & 95.7 & 96.4 & 94.4 & 95.6 \\
        OPT-6.7B & 92.7 & 97.1 & 95.2 & 91.7 & 93.3 \\
        OPT-2.7B & 89.1 & 95.7 & 95.2 & 88.9 & 91.1 \\
        OPT-1.3B & 89.1 & 94.2 & 95.2 & 86.1 & 93.3 \\
        \bottomrule
    \end{tabular}
    \caption{The performance of the models on CNN/DM with FactCC-generated alternative-choices using avg. LL as the scoring function.}
    \label{tab:cnn_dm_factcc_avg_ll}
\end{table*}

\begin{table*}[!ht]
    \small
    \centering
    \begin{tabular}{@{} l c c c c c @{}}
        \toprule
        Model & Date Swap & Entity Swap & Negation & Number Swap & Pronoun \\
       \midrule
        T0-3B & 74.5 & 73.9 & 83.1 & 72.2 & 75.6 \\
        T0 & 78.2 & 72.5 & 88.0 & 63.9 & 66.7 \\
        flan-t5-xl & 76.4 & 73.9 & 79.5 & 75.0 & 64.4 \\
        flan-t5-xxl & 74.5 & 65.2 & 80.7 & 66.7 & 55.6 \\
        t5-lm-adapt-xl & 69.1 & 75.4 & 67.5 & 66.7 & 64.4 \\
        t5-lm-adapt-xxl & 72.7 & 68.1 & 72.3 & 69.4 & 55.6 \\
        GPT-Neo-1.3B & 63.6 & 76.8 & 68.7 & 63.9 & 71.1 \\
        GPT2-XL & 74.5 & 75.4 & 67.5 & 66.7 & 60.0 \\
        GPT-Neo-2.7B & 63.6 & 71.0 & 65.1 & 63.9 & 68.9 \\
        GPTJ-6B & 69.1 & 72.5 & 74.7 & 86.1 & 68.9 \\
        GPT-Neox-20B & 61.8 & 68.1 & 74.7 & 77.8 & 62.2 \\
        BLOOM & 69.1 & 68.1 & 74.7 & 75.0 & 60.0 \\
        BLOOM-7B1 & 74.5 & 73.9 & 74.7 & 69.4 & 75.6 \\
        BLOOM-3B & 74.5 & 76.8 & 65.1 & 72.2 & 66.7 \\
        BLOOM-1B7 & 65.5 & 69.6 & 57.8 & 66.7 & 66.7 \\
        BLOOM-1B1 & 70.9 & 68.1 & 67.5 & 66.7 & 68.9 \\
        OPT-175B & 65.5 & 68.1 & 75.9 & 77.8 & 64.4 \\
        OPT-66B & 61.8 & 68.1 & 71.1 & 69.4 & 68.9 \\
        OPT-30B & 72.7 & 66.7 & 78.3 & 72.2 & 68.9 \\
        OPT-13B & 74.5 & 73.9 & 71.1 & 69.4 & 64.4 \\
        OPT-6.7B & 69.1 & 79.7 & 78.3 & 63.9 & 60.0 \\
        OPT-2.7B & 65.5 & 73.9 & 63.9 & 58.3 & 62.2 \\
        OPT-1.3B & 65.5 & 72.5 & 69.9 & 69.4 & 66.7 \\
        \bottomrule
    \end{tabular}
    \caption{The performance of the models on CNN/DM with FactCC-generated alternative-choices using PMI as the scoring function.}
    \label{tab:cnn_dm_factcc_pmi}
\end{table*}

\begin{table*}[!ht]
    \small
    \centering
    \begin{tabular}{@{} l c c c c c @{}}
        \toprule
        Model & Date Swap & Entity Swap & Negation & Number Swap & Pronoun \\
       \midrule
        T0-3B & 96.4 & 100.0 & 100.0 & 94.4 & 100.0 \\
        T0 & 96.4 & 100.0 & 100.0 & 88.9 & 95.6 \\
        flan-t5-xl & 98.2 & 100.0 & 100.0 & 91.7 & 97.8 \\
        flan-t5-xxl & 96.4 & 98.6 & 98.8 & 88.9 & 97.8 \\
        t5-lm-adapt-xl & 98.2 & 98.6 & 97.6 & 88.9 & 97.8 \\
        t5-lm-adapt-xxl & 96.4 & 100.0 & 100.0 & 94.4 & 100.0 \\
        GPT-Neo-1.3B & 90.9 & 100.0 & 100.0 & 91.7 & 100.0 \\
        GPT2-XL & 92.7 & 97.1 & 98.8 & 94.4 & 97.8 \\
        GPT-Neo-2.7B & 90.9 & 98.6 & 100.0 & 91.7 & 100.0 \\
        GPTJ-6B & 94.5 & 98.6 & 100.0 & 94.4 & 100.0 \\
        GPT-Neox-20B & 94.5 & 100.0 & 100.0 & 91.7 & 100.0 \\
        BLOOM & 96.4 & 98.6 & 100.0 & 94.4 & 100.0 \\
        BLOOM-7B1 & 94.5 & 98.6 & 98.8 & 94.4 & 100.0 \\
        BLOOM-3B & 96.4 & 100.0 & 100.0 & 88.9 & 100.0 \\
        BLOOM-1B7 & 94.5 & 98.6 & 98.8 & 94.4 & 95.6 \\
        BLOOM-1B1 & 94.5 & 100.0 & 98.8 & 91.7 & 97.8 \\
        OPT-175B & 94.5 & 98.6 & 100.0 & 94.4 & 95.6 \\
        OPT-66B & 94.5 & 98.6 & 100.0 & 94.4 & 100.0 \\
        OPT-30B & 94.5 & 98.6 & 100.0 & 94.4 & 100.0 \\
        OPT-13B & 94.5 & 98.6 & 100.0 & 94.4 & 100.0 \\
        OPT-6.7B & 96.4 & 100.0 & 100.0 & 94.4 & 100.0 \\
        OPT-2.7B & 94.5 & 100.0 & 100.0 & 94.4 & 100.0 \\
        OPT-1.3B & 94.5 & 100.0 & 100.0 & 94.4 & 100.0 \\
        \bottomrule
    \end{tabular}
    \caption{The performance of the models on CNN/DM with FactCC-generated alternative-choices using LL as the scoring function.}
    \label{tab:cnn_dm_factcc_ll}
\end{table*}

\section{Accuracies from Factual Model-Generated Summaries}
\label{sec:fmg_scores}

We show the performance of different models on \fc{} model-generated summaries broken down by the model used to generate the summary using different scoring functions on XSum in \cref{tab:xsum_fmg_avg_pmi}, \cref{tab:xsum_fmg_avg_ll}, \cref{tab:xsum_fmg_pmi}, and \cref{tab:xsum_fmg_ll} and on CNN/DM in \cref{tab:cnn_dm_fmg_avg_pmi}, \cref{tab:cnn_dm_fmg_avg_ll}, \cref{tab:cnn_dm_fmg_pmi}, and \cref{tab:cnn_dm_fmg_ll}

\begin{table*}[!ht]
    \small
    \centering
    \begin{tabular}{@{} l c c c c c c c @{}}
        \toprule
        Model & BART- & BART- & BLOOM- & distil- & distil- & PEGASUS & T5- \\
        & base & large & 560m & BART & PEGASUS &  & large\\
       \midrule
        T0-3B & 62.2 & 33.7 & 90.5 & 32.2 & 17.5 & 25.8 & 94.1 \\
        T0 & 64.9 & 18.6 & 85.7 & 23.3 & 14.3 & 29.0 & 76.5 \\
        FLAN-T5-xl & 64.9 & 38.4 & 90.5 & 38.9 & 25.4 & 38.7 & 82.4 \\
        FLAN-T5-xxl & 70.3 & 46.5 & 90.5 & 42.2 & 28.6 & 35.5 & 82.4 \\
        T5-LM-Adapt-xl & 56.8 & 45.3 & 76.2 & 44.4 & 31.7 & 35.5 & 82.4 \\
        T5-LM-Adapt-xxl & 59.5 & 45.3 & 71.4 & 45.6 & 34.9 & 38.7 & 76.5 \\
        GPT-Neo-1.3B & 59.5 & 38.4 & 66.7 & 53.3 & 28.6 & 22.6 & 76.5 \\
        GPT2-XL & 62.2 & 40.7 & 61.9 & 50.0 & 27.0 & 33.9 & 52.9 \\
        GPT-Neo-2.7B & 56.8 & 41.9 & 57.1 & 52.2 & 28.6 & 33.9 & 76.5 \\
        GPTJ-6B & 64.9 & 40.7 & 71.4 & 61.1 & 38.1 & 29.0 & 64.7 \\
        GPT-Neox-20B & 73.0 & 36.0 & 61.9 & 58.9 & 33.3 & 32.3 & 64.7 \\
        BLOOM & 56.8 & 41.9 & 71.4 & 51.1 & 27.0 & 25.8 & 70.6 \\
        BLOOM-7B1 & 56.8 & 34.9 & 52.4 & 50.0 & 30.2 & 27.4 & 70.6 \\
        BLOOM-3B & 64.9 & 30.2 & 57.1 & 50.0 & 23.8 & 32.3 & 64.7 \\
        BLOOM-1B7 & 70.3 & 33.7 & 52.4 & 45.6 & 22.2 & 29.0 & 70.6 \\
        BLOOM-1B1 & 62.2 & 32.6 & 57.1 & 43.3 & 22.2 & 30.6 & 58.8 \\
        OPT-175B & 59.5 & 41.9 & 66.7 & 52.2 & 34.9 & 25.8 & 76.5 \\
        OPT-66B & 75.7 & 38.4 & 52.4 & 57.8 & 31.7 & 22.6 & 70.6 \\
        OPT-30B & 62.2 & 39.5 & 52.4 & 55.6 & 38.1 & 27.4 & 70.6 \\
        OPT-13B & 64.9 & 44.2 & 57.1 & 54.4 & 38.1 & 22.6 & 70.6 \\
        OPT-6.7B & 73.0 & 38.4 & 52.4 & 58.9 & 34.9 & 17.7 & 70.6 \\
        OPT-2.7B & 64.9 & 37.2 & 52.4 & 54.4 & 38.1 & 29.0 & 70.6 \\
        OPT-1.3B & 62.2 & 40.7 & 61.9 & 53.3 & 28.6 & 27.4 & 58.8 \\
        \bottomrule
    \end{tabular}
    \caption{The performance of the models on XSum with \fc{} model-generated alternative-choices using avg. PMI as the scoring function.}
    \label{tab:xsum_fmg_avg_pmi}
\end{table*}

\begin{table*}[!ht]
    \small
    \centering
    \begin{tabular}{@{} l c c c c c c c @{}}
        \toprule
        Model & BART- & BART- & BLOOM- & distil- & distil- & PEGASUS & T5- \\
        & base & large & 560m & BART & PEGASUS &  & large\\
       \midrule
        T0-3B & 27.0 & 2.3 & 95.2 & 3.3 & 7.9 & 3.2 & 52.9 \\
        T0 & 51.4 & 9.3 & 95.2 & 6.7 & 4.8 & 8.1 & 58.8 \\
        FLAN-T5-xl & 27.0 & 2.3 & 95.2 & 2.2 & 7.9 & 8.1 & 52.9 \\
        FLAN-T5-xxl & 37.8 & 5.8 & 95.2 & 4.4 & 4.8 & 4.8 & 52.9 \\
        T5-LM-Adapt-xl & 32.4 & 7.0 & 38.1 & 11.1 & 17.5 & 12.9 & 29.4 \\
        T5-LM-Adapt-xxl & 40.5 & 5.8 & 47.6 & 7.8 & 15.9 & 16.1 & 41.2 \\
        GPT-Neo-1.3B & 40.5 & 7.0 & 42.9 & 16.7 & 6.3 & 11.3 & 41.2 \\
        GPT2-XL & 35.1 & 5.8 & 47.6 & 13.3 & 14.3 & 14.5 & 47.1 \\
        GPT-Neo-2.7B & 35.1 & 10.5 & 38.1 & 18.9 & 9.5 & 12.9 & 41.2 \\
        GPTJ-6B & 51.4 & 9.3 & 52.4 & 17.8 & 9.5 & 8.1 & 47.1 \\
        GPT-Neox-20B & 51.4 & 5.8 & 52.4 & 21.1 & 9.5 & 8.1 & 47.1 \\
        BLOOM & 51.4 & 10.5 & 66.7 & 20.0 & 9.5 & 12.9 & 58.8 \\
        BLOOM-7B1 & 43.2 & 5.8 & 57.1 & 20.0 & 15.9 & 9.7 & 47.1 \\
        BLOOM-3B & 35.1 & 9.3 & 52.4 & 21.1 & 9.5 & 14.5 & 35.3 \\
        BLOOM-1B7 & 32.4 & 10.5 & 47.6 & 22.2 & 15.9 & 9.7 & 35.3 \\
        BLOOM-1B1 & 27.0 & 11.6 & 47.6 & 22.2 & 12.7 & 16.1 & 23.5 \\
        OPT-175B & 56.8 & 7.0 & 66.7 & 20.0 & 11.1 & 9.7 & 47.1 \\
        OPT-66B & 54.1 & 5.8 & 66.7 & 20.0 & 12.7 & 9.7 & 47.1 \\
        OPT-30B & 48.6 & 7.0 & 61.9 & 18.9 & 9.5 & 9.7 & 52.9 \\
        OPT-13B & 51.4 & 5.8 & 61.9 & 17.8 & 7.9 & 9.7 & 58.8 \\
        OPT-6.7B & 51.4 & 4.7 & 47.6 & 15.6 & 12.7 & 12.9 & 58.8 \\
        OPT-2.7B & 45.9 & 4.7 & 47.6 & 18.9 & 12.7 & 11.3 & 41.2 \\
        OPT-1.3B & 43.2 & 5.8 & 52.4 & 17.8 & 12.7 & 9.7 & 41.2 \\
        \bottomrule
    \end{tabular}
    \caption{The performance of the models on XSum with \fc{} model-generated alternative-choices using avg. LL as the scoring function.}
    \label{tab:xsum_fmg_avg_ll}
\end{table*}

\begin{table*}[!ht]
    \small
    \centering
    \begin{tabular}{@{} l c c c c c c c @{}}
        \toprule
        Model & BART- & BART- & BLOOM- & distil- & distil- & PEGASUS & T5- \\
        & base & large & 560m & BART & PEGASUS &  & large\\
       \midrule
        T0-3B & 64.9 & 27.9 & 66.7 & 34.4 & 38.1 & 45.2 & 76.5 \\
        T0 & 64.9 & 18.6 & 81.0 & 22.2 & 22.2 & 32.3 & 82.4 \\
        FLAN-T5-xl & 59.5 & 39.5 & 66.7 & 44.4 & 47.6 & 48.4 & 58.8 \\
        FLAN-T5-xxl & 59.5 & 40.7 & 57.1 & 40.0 & 49.2 & 46.8 & 64.7 \\
        T5-LM-Adapt-xl & 56.8 & 40.7 & 38.1 & 48.9 & 50.8 & 51.6 & 64.7 \\
        T5-LM-Adapt-xxl & 59.5 & 41.9 & 42.9 & 43.3 & 47.6 & 51.6 & 58.8 \\
        GPT-Neo-1.3B & 67.6 & 36.0 & 4.8 & 54.4 & 42.9 & 35.5 & 58.8 \\
        GPT2-XL & 67.6 & 38.4 & 28.6 & 53.3 & 49.2 & 46.8 & 52.9 \\
        GPT-Neo-2.7B & 64.9 & 37.2 & 9.5 & 56.7 & 46.0 & 43.5 & 58.8 \\
        GPTJ-6B & 70.3 & 40.7 & 9.5 & 62.2 & 55.6 & 48.4 & 58.8 \\
        GPT-Neox-20B & 73.0 & 31.4 & 19.0 & 55.6 & 46.0 & 45.2 & 58.8 \\
        BLOOM & 67.6 & 45.3 & 14.3 & 44.4 & 41.3 & 40.3 & 70.6 \\
        BLOOM-7B1 & 62.2 & 40.7 & 9.5 & 53.3 & 42.9 & 40.3 & 64.7 \\
        BLOOM-3B & 73.0 & 34.9 & 19.0 & 54.4 & 36.5 & 48.4 & 64.7 \\
        BLOOM-1B7 & 62.2 & 37.2 & 14.3 & 43.3 & 39.7 & 48.4 & 52.9 \\
        BLOOM-1B1 & 62.2 & 32.6 & 9.5 & 46.7 & 38.1 & 46.8 & 52.9 \\
        OPT-175B & 67.6 & 40.7 & 9.5 & 54.4 & 49.2 & 38.7 & 70.6 \\
        OPT-66B & 75.7 & 38.4 & 4.8 & 54.4 & 52.4 & 37.1 & 70.6 \\
        OPT-30B & 67.6 & 43.0 & 14.3 & 52.2 & 46.0 & 38.7 & 58.8 \\
        OPT-13B & 64.9 & 43.0 & 9.5 & 53.3 & 50.8 & 41.9 & 64.7 \\
        OPT-6.7B & 73.0 & 38.4 & 4.8 & 58.9 & 52.4 & 37.1 & 52.9 \\
        OPT-2.7B & 73.0 & 40.7 & 9.5 & 57.8 & 52.4 & 40.3 & 58.8 \\
        OPT-1.3B & 64.9 & 43.0 & 4.8 & 52.2 & 44.4 & 43.5 & 47.1 \\
        \bottomrule
    \end{tabular}
    \caption{The performance of the models on XSum with \fc{} model-generated alternative-choices using PMI as the scoring function.}
    \label{tab:xsum_fmg_pmi}
\end{table*}

\begin{table*}[!ht]
    \small  
    \centering
    \begin{tabular}{@{} l c c c c c c c @{}}
        \toprule
        Model & BART- & BART- & BLOOM- & distil- & distil- & PEGASUS & T5- \\
        & base & large & 560m & BART & PEGASUS &  & large\\
       \midrule
        T0-3B & 21.6 & 4.7 & 100.0 & 5.6 & 6.3 & 4.8 & 47.1 \\
        T0 & 48.6 & 9.3 & 100.0 & 10.0 & 6.3 & 9.7 & 64.7 \\
        FLAN-T5-xl & 27.0 & 7.0 & 100.0 & 4.4 & 6.3 & 11.3 & 52.9 \\
        FLAN-T5-xxl & 32.4 & 7.0 & 100.0 & 4.4 & 3.2 & 12.9 & 47.1 \\
        T5-LM-Adapt-xl & 32.4 & 12.8 & 95.2 & 15.6 & 14.3 & 11.3 & 47.1 \\
        T5-LM-Adapt-xxl & 32.4 & 10.5 & 90.5 & 11.1 & 11.1 & 11.3 & 58.8 \\
        GPT-Neo-1.3B & 37.8 & 9.3 & 85.7 & 23.3 & 6.3 & 11.3 & 35.3 \\
        GPT2-XL & 32.4 & 8.1 & 85.7 & 16.7 & 9.5 & 14.5 & 52.9 \\
        GPT-Neo-2.7B & 37.8 & 9.3 & 85.7 & 23.3 & 7.9 & 11.3 & 47.1 \\
        GPTJ-6B & 35.1 & 7.0 & 95.2 & 22.2 & 11.1 & 8.1 & 52.9 \\
        GPT-Neox-20B & 51.4 & 10.5 & 95.2 & 26.7 & 9.5 & 9.7 & 58.8 \\
        BLOOM & 40.5 & 12.8 & 95.2 & 17.8 & 7.9 & 9.7 & 64.7 \\
        BLOOM-7B1 & 40.5 & 9.3 & 90.5 & 23.3 & 9.5 & 11.3 & 52.9 \\
        BLOOM-3B & 37.8 & 10.5 & 90.5 & 23.3 & 11.1 & 12.9 & 41.2 \\
        BLOOM-1B7 & 40.5 & 12.8 & 85.7 & 25.6 & 11.1 & 12.9 & 41.2 \\
        BLOOM-1B1 & 32.4 & 16.3 & 81.0 & 23.3 & 9.5 & 12.9 & 47.1 \\
        OPT-175B & 51.4 & 9.3 & 95.2 & 24.4 & 6.3 & 12.9 & 64.7 \\
        OPT-66B & 43.2 & 11.6 & 95.2 & 21.1 & 7.9 & 11.3 & 64.7 \\
        OPT-30B & 45.9 & 10.5 & 95.2 & 23.3 & 4.8 & 12.9 & 64.7 \\
        OPT-13B & 48.6 & 9.3 & 95.2 & 20.0 & 6.3 & 9.7 & 64.7 \\
        OPT-6.7B & 45.9 & 9.3 & 95.2 & 23.3 & 9.5 & 11.3 & 64.7 \\
        OPT-2.7B & 37.8 & 12.8 & 95.2 & 20.0 & 7.9 & 11.3 & 58.8 \\
        OPT-1.3B & 37.8 & 12.8 & 95.2 & 20.0 & 4.8 & 9.7 & 47.1 \\
        \bottomrule
    \end{tabular}
    \caption{The performance of the models on XSum with \fc{} model-generated alternative-choices using LL as the scoring function.}
    \label{tab:xsum_fmg_ll}
\end{table*}

\begin{table*}[!ht]
    \centering
    \footnotesize
    \begin{tabular}{@{} l c c c c c c c c c c c c c c c @{}}
        \toprule
        Model & B & BL & HG & L & MS & MI & NS & OD & O & PB & PT & R & RE & T & TS \\
       \midrule
        T0-3B & 1.4 & 3.9 & 1.3 & 2.1 & 5.1 & 4.5 & 23.7 & 39.3 & 8.7 & 3.4 & 0.0 & 23.2 & 2.6 & 4.7 & 3.7 \\
        T0 & 2.7 & 3.9 & 0.0 & 1.1 & 2.5 & 6.1 & 10.5 & 21.4 & 4.3 & 1.1 & 1.4 & 8.7 & 3.9 & 4.7 & 7.4 \\
        FLAN-T5-xl & 1.4 & 3.9 & 1.3 & 0.0 & 3.8 & 3.0 & 25.0 & 28.6 & 0.0 & 2.3 & 1.4 & 23.2 & 5.3 & 6.2 & 5.6 \\
        FLAN-T5-xxl & 2.7 & 2.6 & 1.3 & 1.1 & 2.5 & 3.0 & 14.5 & 35.7 & 0.0 & 0.0 & 1.4 & 15.9 & 6.6 & 4.7 & 1.9 \\
        T5-LM-Adapt-xl & 5.4 & 2.6 & 0.0 & 0.0 & 0.0 & 3.0 & 18.4 & 35.7 & 0.0 & 0.0 & 0.0 & 20.3 & 2.6 & 3.1 & 1.9 \\
        T5-LM-Adapt-xxl & 5.4 & 5.2 & 2.6 & 1.1 & 5.1 & 6.1 & 14.5 & 28.6 & 0.0 & 2.3 & 1.4 & 17.4 & 5.3 & 6.2 & 1.9 \\
        GPT-Neo-1.3B & 1.4 & 1.3 & 0.0 & 1.1 & 3.8 & 4.5 & 35.5 & 32.1 & 2.2 & 1.1 & 2.7 & 20.3 & 2.6 & 3.1 & 0.0 \\
        GPT2-XL & 1.4 & 2.6 & 2.6 & 1.1 & 2.5 & 6.1 & 44.7 & 14.3 & 0.0 & 2.3 & 2.7 & 40.6 & 2.6 & 0.0 & 1.9 \\
        GPT-Neo-2.7B & 4.1 & 3.9 & 3.8 & 1.1 & 6.3 & 3.0 & 31.6 & 28.6 & 2.2 & 2.3 & 2.7 & 24.6 & 6.6 & 6.2 & 3.7 \\
        GPTJ-6B & 4.1 & 5.2 & 5.1 & 2.1 & 5.1 & 6.1 & 25.0 & 14.3 & 2.2 & 3.4 & 6.8 & 20.3 & 6.6 & 6.2 & 3.7 \\
        GPT-Neox-20B & 5.4 & 6.5 & 6.4 & 2.1 & 8.9 & 7.6 & 23.7 & 14.3 & 4.3 & 5.7 & 6.8 & 23.2 & 7.9 & 6.2 & 3.7 \\
        BLOOM & 5.4 & 5.2 & 7.7 & 5.3 & 11.4 & 9.1 & 28.9 & 17.9 & 4.3 & 6.8 & 8.2 & 26.1 & 14.5 & 10.9 & 3.7 \\
        BLOOM-7B1 & 4.1 & 5.2 & 6.4 & 5.3 & 5.1 & 9.1 & 27.6 & 25.0 & 6.5 & 5.7 & 8.2 & 24.6 & 7.9 & 10.9 & 5.6 \\
        BLOOM-3B & 5.4 & 5.2 & 3.8 & 3.2 & 3.8 & 4.5 & 28.9 & 28.6 & 2.2 & 4.5 & 4.1 & 20.3 & 5.3 & 7.8 & 3.7 \\
        BLOOM-1B7 & 2.7 & 2.6 & 2.6 & 1.1 & 3.8 & 3.0 & 27.6 & 32.1 & 2.2 & 2.3 & 2.7 & 23.2 & 5.3 & 4.7 & 1.9 \\
        BLOOM-1B1 & 2.7 & 2.6 & 1.3 & 0.0 & 5.1 & 4.5 & 31.6 & 32.1 & 2.2 & 1.1 & 4.1 & 27.5 & 7.9 & 4.7 & 0.0 \\
        OPT-175B & 10.8 & 11.7 & 11.5 & 5.3 & 10.1 & 10.6 & 30.3 & 14.3 & 4.3 & 8.0 & 9.6 & 20.3 & 13.2 & 10.9 & 7.4 \\
        OPT-66B & 9.5 & 9.1 & 9.0 & 3.2 & 8.9 & 6.1 & 19.7 & 10.7 & 4.3 & 5.7 & 8.2 & 15.9 & 9.2 & 7.8 & 5.6 \\
        OPT-30B & 14.9 & 10.4 & 9.0 & 4.2 & 10.1 & 10.6 & 25.0 & 10.7 & 6.5 & 9.1 & 9.6 & 17.4 & 11.8 & 9.4 & 7.4 \\
        OPT-13B & 6.8 & 6.5 & 5.1 & 2.1 & 6.3 & 7.6 & 23.7 & 14.3 & 2.2 & 4.5 & 8.2 & 20.3 & 7.9 & 7.8 & 3.7 \\
        OPT-6.7B & 8.1 & 7.8 & 9.0 & 4.2 & 7.6 & 9.1 & 25.0 & 17.9 & 6.5 & 6.8 & 8.2 & 21.7 & 10.5 & 9.4 & 5.6 \\
        OPT-2.7B & 6.8 & 5.2 & 5.1 & 1.1 & 3.8 & 6.1 & 26.3 & 17.9 & 4.3 & 4.5 & 5.5 & 20.3 & 6.6 & 7.8 & 1.9 \\
        OPT-1.3B & 9.5 & 5.2 & 3.8 & 1.1 & 3.8 & 6.1 & 23.7 & 17.9 & 2.2 & 2.3 & 4.1 & 15.9 & 5.3 & 6.2 & 1.9 \\
        \bottomrule
    \end{tabular}
    \caption{The performance of the models on CNN/DM with \fc{} model-generated alternative-choices using avg. PMI as the scoring function. The models are   BanditSumm (B), BERT\_LSTM\_PN\_RL (BL), Heter-Graph (HG), Lead3 (L), MatchSumm (MS), MI-unsup (MI), NeuSumm (NS), Oracle (discourse) (OD), Oracle (O), Pacsum (bert) (PB), Pacsum (tfidf) (PT), Refresh (R), RNN\_Ext\_RL (RE), Textrank (T), Textrank (st) (TS) \\}
    \label{tab:cnn_dm_fmg_avg_pmi}
\end{table*}

\begin{table*}[!ht]
    \centering
    \footnotesize
    \begin{tabular}{@{} l c c c c c c c c c c c c c c c @{}}
        \toprule
        Model & B & BL & HG & L & MS & MI & NS & OD & O & PB & PT & R & RE & T & TS \\
       \midrule
        T0-3B & 1.4 & 0.0 & 0.0 & 1.1 & 1.3 & 1.5 & 13.2 & 14.3 & 2.2 & 0.0 & 1.4 & 8.7 & 5.3 & 3.1 & 3.7 \\
        T0 & 1.4 & 0.0 & 0.0 & 1.1 & 0.0 & 3.0 & 3.9 & 10.7 & 4.3 & 0.0 & 1.4 & 1.4 & 6.6 & 3.1 & 3.7 \\
        FLAN-T5-xl & 1.4 & 0.0 & 0.0 & 0.0 & 0.0 & 0.0 & 5.3 & 0.0 & 4.3 & 0.0 & 0.0 & 5.8 & 0.0 & 4.7 & 3.7 \\
        FLAN-T5-xxl & 0.0 & 0.0 & 0.0 & 0.0 & 0.0 & 1.5 & 1.3 & 3.6 & 2.2 & 0.0 & 0.0 & 1.4 & 0.0 & 3.1 & 3.7 \\
        T5-LM-Adapt-xl & 0.0 & 0.0 & 0.0 & 0.0 & 0.0 & 0.0 & 5.3 & 7.1 & 2.2 & 0.0 & 1.4 & 5.8 & 2.6 & 3.1 & 1.9 \\
        T5-LM-Adapt-xxl & 1.4 & 0.0 & 0.0 & 0.0 & 0.0 & 1.5 & 1.3 & 3.6 & 2.2 & 0.0 & 1.4 & 1.4 & 5.3 & 3.1 & 0.0 \\
        GPT-Neo-1.3B & 0.0 & 0.0 & 0.0 & 0.0 & 0.0 & 0.0 & 2.6 & 0.0 & 2.2 & 0.0 & 0.0 & 4.3 & 0.0 & 1.6 & 0.0 \\
        GPT2-XL & 0.0 & 0.0 & 0.0 & 0.0 & 0.0 & 1.5 & 3.9 & 0.0 & 2.2 & 0.0 & 0.0 & 4.3 & 0.0 & 1.6 & 0.0 \\
        GPT-Neo-2.7B & 0.0 & 0.0 & 0.0 & 0.0 & 0.0 & 0.0 & 3.9 & 0.0 & 2.2 & 0.0 & 0.0 & 4.3 & 0.0 & 1.6 & 0.0 \\
        GPTJ-6B & 0.0 & 0.0 & 0.0 & 0.0 & 0.0 & 0.0 & 2.6 & 0.0 & 2.2 & 0.0 & 0.0 & 1.4 & 0.0 & 1.6 & 0.0 \\
        GPT-Neox-20B & 0.0 & 0.0 & 0.0 & 0.0 & 0.0 & 0.0 & 0.0 & 0.0 & 2.2 & 0.0 & 0.0 & 0.0 & 0.0 & 1.6 & 0.0 \\
        BLOOM & 0.0 & 0.0 & 0.0 & 0.0 & 0.0 & 0.0 & 1.3 & 0.0 & 2.2 & 0.0 & 0.0 & 0.0 & 0.0 & 1.6 & 0.0 \\
        BLOOM-7B1 & 0.0 & 0.0 & 0.0 & 0.0 & 0.0 & 1.5 & 3.9 & 0.0 & 2.2 & 0.0 & 0.0 & 4.3 & 0.0 & 1.6 & 1.9 \\
        BLOOM-3B & 0.0 & 0.0 & 0.0 & 0.0 & 0.0 & 1.5 & 5.3 & 0.0 & 2.2 & 0.0 & 0.0 & 5.8 & 0.0 & 1.6 & 1.9 \\
        BLOOM-1B7 & 1.4 & 0.0 & 0.0 & 0.0 & 0.0 & 1.5 & 2.6 & 3.6 & 2.2 & 0.0 & 0.0 & 4.3 & 0.0 & 0.0 & 0.0 \\
        BLOOM-1B1 & 2.7 & 1.3 & 0.0 & 1.1 & 0.0 & 1.5 & 2.6 & 0.0 & 2.2 & 0.0 & 0.0 & 5.8 & 0.0 & 1.6 & 1.9 \\
        OPT-175B & 1.4 & 0.0 & 0.0 & 0.0 & 0.0 & 1.5 & 1.3 & 0.0 & 2.2 & 0.0 & 0.0 & 1.4 & 0.0 & 1.6 & 0.0 \\
        OPT-66B & 1.4 & 0.0 & 0.0 & 0.0 & 0.0 & 1.5 & 3.9 & 0.0 & 2.2 & 0.0 & 0.0 & 2.9 & 0.0 & 1.6 & 0.0 \\
        OPT-30B & 0.0 & 0.0 & 0.0 & 0.0 & 0.0 & 1.5 & 3.9 & 0.0 & 2.2 & 0.0 & 0.0 & 2.9 & 0.0 & 1.6 & 0.0 \\
        OPT-13B & 0.0 & 0.0 & 0.0 & 0.0 & 0.0 & 1.5 & 5.3 & 0.0 & 2.2 & 0.0 & 0.0 & 2.9 & 0.0 & 1.6 & 0.0 \\
        OPT-6.7B & 0.0 & 0.0 & 0.0 & 0.0 & 0.0 & 1.5 & 6.6 & 0.0 & 2.2 & 0.0 & 0.0 & 1.4 & 0.0 & 1.6 & 0.0 \\
        OPT-2.7B & 1.4 & 0.0 & 0.0 & 0.0 & 0.0 & 1.5 & 6.6 & 0.0 & 2.2 & 0.0 & 0.0 & 4.3 & 0.0 & 1.6 & 0.0 \\
        OPT-1.3B & 1.4 & 0.0 & 0.0 & 0.0 & 0.0 & 1.5 & 6.6 & 0.0 & 2.2 & 0.0 & 0.0 & 4.3 & 0.0 & 1.6 & 0.0 \\
        \bottomrule
    \end{tabular}
    \caption{The performance of the models on CNN/DM with \fc{} model-generated alternative-choices using avg. LL as the scoring function. The models are   BanditSumm (B), BERT\_LSTM\_PN\_RL (BL), Heter-Graph (HG), Lead3 (L), MatchSumm (MS), MI-unsup (MI), NeuSumm (NS), Oracle (discourse) (OD), Oracle (O), Pacsum (bert) (PB), Pacsum (tfidf) (PT), Refresh (R), RNN\_Ext\_RL (RE), Textrank (T), Textrank (st) (TS) \\}
    \label{tab:cnn_dm_fmg_avg_ll}
\end{table*}

\begin{table*}[!ht]
    \centering
    \footnotesize
    \begin{tabular}{@{} l c c c c c c c c c c c c c c c @{}}
        \toprule
        Model & B & BL & HG & L & MS & MI & NS & OD & O & PB & PT & R & RE & T & TS \\
       \midrule
        T0-3B & 1.4 & 1.3 & 0.0 & 0.0 & 0.0 & 0.0 & 1.3 & 21.4 & 6.5 & 0.0 & 0.0 & 1.4 & 0.0 & 4.7 & 1.9 \\
        T0 & 1.4 & 0.0 & 0.0 & 0.0 & 0.0 & 0.0 & 0.0 & 14.3 & 6.5 & 0.0 & 0.0 & 0.0 & 0.0 & 4.7 & 1.9 \\
        FLAN-T5-xl & 0.0 & 1.3 & 0.0 & 0.0 & 0.0 & 0.0 & 1.3 & 10.7 & 4.3 & 0.0 & 0.0 & 0.0 & 0.0 & 4.7 & 1.9 \\
        FLAN-T5-xxl & 1.4 & 0.0 & 0.0 & 0.0 & 0.0 & 0.0 & 0.0 & 14.3 & 4.3 & 0.0 & 0.0 & 0.0 & 0.0 & 3.1 & 1.9 \\
        T5-LM-Adapt-xl & 0.0 & 0.0 & 0.0 & 0.0 & 0.0 & 0.0 & 0.0 & 17.9 & 4.3 & 0.0 & 0.0 & 0.0 & 0.0 & 4.7 & 1.9 \\
        T5-LM-Adapt-xxl & 0.0 & 0.0 & 0.0 & 0.0 & 0.0 & 0.0 & 0.0 & 14.3 & 4.3 & 0.0 & 0.0 & 0.0 & 0.0 & 3.1 & 1.9 \\
        GPT-Neo-1.3B & 0.0 & 0.0 & 0.0 & 0.0 & 0.0 & 0.0 & 0.0 & 3.6 & 4.3 & 0.0 & 0.0 & 0.0 & 0.0 & 1.6 & 1.9 \\
        GPT2-XL & 0.0 & 0.0 & 0.0 & 0.0 & 0.0 & 0.0 & 0.0 & 3.6 & 2.2 & 0.0 & 0.0 & 0.0 & 0.0 & 1.6 & 1.9 \\
        GPT-Neo-2.7B & 0.0 & 0.0 & 0.0 & 0.0 & 0.0 & 0.0 & 0.0 & 3.6 & 4.3 & 0.0 & 0.0 & 0.0 & 0.0 & 0.0 & 1.9 \\
        GPTJ-6B & 0.0 & 0.0 & 0.0 & 0.0 & 0.0 & 0.0 & 0.0 & 3.6 & 4.3 & 0.0 & 0.0 & 0.0 & 0.0 & 1.6 & 1.9 \\
        GPT-Neox-20B & 0.0 & 0.0 & 0.0 & 0.0 & 0.0 & 0.0 & 0.0 & 3.6 & 4.3 & 0.0 & 0.0 & 0.0 & 0.0 & 0.0 & 1.9 \\
        BLOOM & 0.0 & 0.0 & 0.0 & 0.0 & 0.0 & 0.0 & 0.0 & 3.6 & 4.3 & 0.0 & 0.0 & 0.0 & 0.0 & 1.6 & 1.9 \\
        BLOOM-7B1 & 0.0 & 0.0 & 0.0 & 0.0 & 0.0 & 0.0 & 0.0 & 3.6 & 4.3 & 0.0 & 0.0 & 0.0 & 0.0 & 1.6 & 1.9 \\
        BLOOM-3B & 0.0 & 0.0 & 0.0 & 0.0 & 0.0 & 0.0 & 0.0 & 3.6 & 4.3 & 0.0 & 0.0 & 0.0 & 0.0 & 1.6 & 1.9 \\
        BLOOM-1B7 & 0.0 & 0.0 & 0.0 & 0.0 & 0.0 & 0.0 & 0.0 & 3.6 & 4.3 & 0.0 & 0.0 & 0.0 & 0.0 & 1.6 & 1.9 \\
        BLOOM-1B1 & 0.0 & 0.0 & 0.0 & 0.0 & 0.0 & 0.0 & 0.0 & 3.6 & 4.3 & 0.0 & 0.0 & 0.0 & 0.0 & 1.6 & 1.9 \\
        OPT-175B & 1.4 & 0.0 & 0.0 & 0.0 & 0.0 & 0.0 & 0.0 & 3.6 & 4.3 & 0.0 & 0.0 & 0.0 & 0.0 & 3.1 & 1.9 \\
        OPT-66B & 0.0 & 0.0 & 0.0 & 0.0 & 0.0 & 0.0 & 0.0 & 3.6 & 4.3 & 0.0 & 0.0 & 0.0 & 0.0 & 1.6 & 1.9 \\
        OPT-30B & 0.0 & 0.0 & 0.0 & 0.0 & 0.0 & 0.0 & 0.0 & 3.6 & 4.3 & 0.0 & 0.0 & 0.0 & 0.0 & 3.1 & 1.9 \\
        OPT-13B & 0.0 & 0.0 & 0.0 & 0.0 & 0.0 & 0.0 & 0.0 & 3.6 & 4.3 & 0.0 & 0.0 & 0.0 & 0.0 & 3.1 & 1.9 \\
        OPT-6.7B & 0.0 & 0.0 & 0.0 & 0.0 & 0.0 & 0.0 & 0.0 & 3.6 & 4.3 & 0.0 & 0.0 & 0.0 & 0.0 & 1.6 & 1.9 \\
        OPT-2.7B & 0.0 & 0.0 & 0.0 & 0.0 & 0.0 & 0.0 & 0.0 & 3.6 & 4.3 & 0.0 & 0.0 & 0.0 & 0.0 & 1.6 & 1.9 \\
        OPT-1.3B & 0.0 & 0.0 & 0.0 & 0.0 & 0.0 & 0.0 & 0.0 & 3.6 & 4.3 & 0.0 & 0.0 & 0.0 & 0.0 & 1.6 & 1.9 \\
        \bottomrule
    \end{tabular}
    \caption{The performance of the models on CNN/DM with \fc{} model-generated alternative-choices using PMI as the scoring function. The models are   BanditSumm (B), BERT\_LSTM\_PN\_RL (BL), Heter-Graph (HG), Lead3 (L), MatchSumm (MS), MI-unsup (MI), NeuSumm (NS), Oracle (discourse) (OD), Oracle (O), Pacsum (bert) (PB), Pacsum (tfidf) (PT), Refresh (R), RNN\_Ext\_RL (RE), Textrank (T), Textrank (st) (TS) \\}
    \label{tab:cnn_dm_fmg_pmi}
\end{table*}

\begin{table*}[!ht]
    \centering
    \tiny
    \begin{tabular}{@{} l c c c c c c c c c c c c c c c @{}}
        \toprule
        Model & B & BL & HG & L & MS & MI & NS & OD & O & PB & PT & R & RE & T & TS \\
       \midrule
        T0-3B & 31.1 & 24.7 & 42.3 & 25.3 & 44.3 & 47.0 & 98.7 & 75.0 & 21.7 & 30.7 & 35.6 & 88.4 & 64.5 & 31.2 & 29.6 \\
        T0 & 35.1 & 20.8 & 26.9 & 13.7 & 26.6 & 45.5 & 85.5 & 57.1 & 23.9 & 23.9 & 28.8 & 76.8 & 47.4 & 32.8 & 35.2 \\
        FLAN-T5-xl & 31.1 & 26.0 & 37.2 & 21.1 & 32.9 & 48.5 & 94.7 & 53.6 & 19.6 & 30.7 & 28.8 & 91.3 & 56.6 & 35.9 & 33.3 \\
        FLAN-T5-xxl & 20.3 & 11.7 & 16.7 & 8.4 & 15.2 & 36.4 & 76.3 & 46.4 & 13.0 & 12.5 & 17.8 & 55.1 & 38.2 & 15.6 & 20.4 \\
        T5-LM-Adapt-xl & 52.7 & 41.6 & 28.2 & 12.6 & 22.8 & 60.6 & 94.7 & 57.1 & 17.4 & 21.6 & 23.3 & 82.6 & 48.7 & 20.3 & 22.2 \\
        T5-LM-Adapt-xxl & 47.3 & 36.4 & 14.1 & 6.3 & 13.9 & 57.6 & 82.9 & 32.1 & 10.9 & 11.4 & 15.1 & 68.1 & 40.8 & 15.6 & 22.2 \\
        GPT-Neo-1.3B & 31.1 & 24.7 & 9.0 & 2.1 & 8.9 & 36.4 & 85.5 & 25.0 & 2.2 & 9.1 & 17.8 & 63.8 & 19.7 & 14.1 & 16.7 \\
        GPT2-XL & 35.1 & 27.3 & 7.7 & 7.4 & 6.3 & 50.0 & 89.5 & 25.0 & 0.0 & 11.4 & 16.4 & 69.6 & 27.6 & 12.5 & 16.7 \\
        GPT-Neo-2.7B & 35.1 & 28.6 & 5.1 & 1.1 & 7.6 & 43.9 & 85.5 & 21.4 & 0.0 & 8.0 & 15.1 & 55.1 & 26.3 & 10.9 & 16.7 \\
        GPTJ-6B & 27.0 & 23.4 & 9.0 & 1.1 & 8.9 & 39.4 & 73.7 & 17.9 & 2.2 & 6.8 & 12.3 & 44.9 & 21.1 & 12.5 & 14.8 \\
        GPT-Neox-20B & 28.4 & 24.7 & 10.3 & 4.2 & 10.1 & 31.8 & 72.4 & 21.4 & 4.3 & 9.1 & 13.7 & 44.9 & 27.6 & 18.8 & 16.7 \\
        BLOOM & 18.9 & 14.3 & 7.7 & 0.0 & 5.1 & 27.3 & 57.9 & 21.4 & 0.0 & 4.5 & 12.3 & 36.2 & 25.0 & 9.4 & 14.8 \\
        BLOOM-7B1 & 31.1 & 22.1 & 6.4 & 3.2 & 6.3 & 39.4 & 80.3 & 21.4 & 0.0 & 9.1 & 13.7 & 46.4 & 23.7 & 12.5 & 14.8 \\
        BLOOM-3B & 41.9 & 31.2 & 9.0 & 3.2 & 10.1 & 45.5 & 80.3 & 25.0 & 0.0 & 8.0 & 13.7 & 60.9 & 23.7 & 10.9 & 14.8 \\
        BLOOM-1B7 & 36.5 & 28.6 & 7.7 & 2.1 & 7.6 & 43.9 & 82.9 & 28.6 & 2.2 & 4.5 & 15.1 & 58.0 & 21.1 & 6.2 & 9.3 \\
        BLOOM-1B1 & 36.5 & 28.6 & 7.7 & 5.3 & 10.1 & 47.0 & 84.2 & 25.0 & 2.2 & 9.1 & 16.4 & 65.2 & 26.3 & 14.1 & 14.8 \\
        OPT-175B & 45.9 & 33.8 & 11.5 & 1.1 & 8.9 & 48.5 & 78.9 & 25.0 & 2.2 & 10.2 & 12.3 & 55.1 & 25.0 & 14.1 & 16.7 \\
        OPT-66B & 44.6 & 33.8 & 10.3 & 4.2 & 6.3 & 48.5 & 82.9 & 21.4 & 4.3 & 12.5 & 16.4 & 49.3 & 28.9 & 17.2 & 16.7 \\
        OPT-30B & 44.6 & 31.2 & 7.7 & 3.2 & 6.3 & 47.0 & 81.6 & 21.4 & 2.2 & 9.1 & 15.1 & 56.5 & 22.4 & 14.1 & 16.7 \\
        OPT-13B & 47.3 & 33.8 & 10.3 & 2.1 & 8.9 & 48.5 & 86.8 & 25.0 & 8.7 & 11.4 & 16.4 & 59.4 & 28.9 & 17.2 & 18.5 \\
        OPT-6.7B & 44.6 & 33.8 & 11.5 & 4.2 & 8.9 & 54.5 & 89.5 & 28.6 & 6.5 & 12.5 & 19.2 & 62.3 & 27.6 & 20.3 & 20.4 \\
        OPT-2.7B & 45.9 & 36.4 & 14.1 & 3.2 & 8.9 & 50.0 & 86.8 & 21.4 & 6.5 & 14.8 & 19.2 & 63.8 & 31.6 & 17.2 & 20.4 \\
        OPT-1.3B & 45.9 & 40.3 & 10.3 & 3.2 & 8.9 & 50.0 & 85.5 & 17.9 & 2.2 & 12.5 & 16.4 & 63.8 & 21.1 & 15.6 & 18.5 \\
        \bottomrule
    \end{tabular}
    \caption{The performance of the models on CNN/DM with \fc{} model-generated alternative-choices using LL as the scoring function. The models are   BanditSumm (B), BERT\_LSTM\_PN\_RL (BL), Heter-Graph (HG), Lead3 (L), MatchSumm (MS), MI-unsup (MI), NeuSumm (NS), Oracle (discourse) (OD), Oracle (O), Pacsum (bert) (PB), Pacsum (tfidf) (PT), Refresh (R), RNN\_Ext\_RL (RE), Textrank (T), Textrank (st) (TS) \\}
    \label{tab:cnn_dm_fmg_ll}
\end{table*}

\section{Accuracies from \benchmark Summaries}
\label{sec:nfmg_scores}
We show the performance of different models on \benchmark broken down by the model used to generate the summary using different scoring functions for XSum in \cref{tab:xsum_nfmg_avg_pmi}, \cref{tab:xsum_nfmg_avg_ll}, \cref{tab:xsum_nfmg_pmi}, and \cref{tab:xsum_nfmg_ll} and for CNN/DM in \cref{tab:cnn_dm_nfmg_avg_pmi}, \cref{tab:cnn_dm_nfmg_avg_ll}, \cref{tab:cnn_dm_nfmg_pmi}, and \cref{tab:cnn_dm_nfmg_ll}.

\begin{table*}[!ht]
    \small
    \centering
    \begin{tabular}{@{} l c c c c c c c @{}}
        \toprule
        Model & BART- & BART- & BLOOM- & distil- & distil- & PEGASUS & T5- \\
        & base & large & 560m & BART & PEGASUS &  & large\\
       \midrule
        T0-3B & 61.1 & 37.9 & 96.0 & 35.1 & 38.7 & 30.6 & 94.0 \\
        T0 & 55.7 & 19.6 & 91.0 & 20.2 & 19.7 & 15.1 & 92.5 \\
        FLAN-T5-xl & 64.1 & 40.8 & 98.7 & 38.3 & 40.7 & 34.5 & 92.8 \\
        FLAN-T5-xxl & 67.8 & 47.6 & 99.0 & 42.9 & 44.6 & 41.8 & 93.2 \\
        T5-LM-Adapt-xl & 66.5 & 60.9 & 90.8 & 57.1 & 61.1 & 53.4 & 86.3 \\
        T5-LM-Adapt-xxl & 70.6 & 61.6 & 95.4 & 56.3 & 59.0 & 53.0 & 87.0 \\
        GPT-Neo-1.3B & 71.3 & 67.9 & 79.9 & 72.4 & 64.8 & 66.2 & 80.3 \\
        GPT2-XL & 67.8 & 63.5 & 84.7 & 64.4 & 61.6 & 60.3 & 78.9 \\
        GPT-Neo-2.7B & 71.9 & 65.9 & 87.0 & 67.3 & 65.4 & 64.8 & 81.2 \\
        GPTJ-6B & 78.6 & 69.3 & 91.8 & 71.5 & 64.5 & 64.8 & 84.1 \\
        GPT-Neox-20B & 76.5 & 64.7 & 89.3 & 70.5 & 64.1 & 61.9 & 83.6 \\
        BLOOM & 72.1 & 65.0 & 92.7 & 65.1 & 62.9 & 59.6 & 85.1 \\
        BLOOM-7B1 & 71.5 & 64.7 & 86.4 & 66.8 & 63.6 & 63.7 & 83.0 \\
        BLOOM-3B & 70.8 & 68.8 & 85.7 & 68.5 & 65.0 & 66.2 & 80.7 \\
        BLOOM-1B7 & 68.3 & 67.1 & 82.6 & 68.5 & 65.0 & 61.6 & 78.3 \\
        BLOOM-1B1 & 66.5 & 63.5 & 80.7 & 66.1 & 65.4 & 63.2 & 73.9 \\
        OPT-175B & 78.8 & 66.4 & 91.0 & 67.8 & 65.2 & 63.2 & 89.4 \\
        OPT-66B & 76.7 & 66.7 & 88.5 & 67.6 & 64.5 & 61.6 & 88.0 \\
        OPT-30B & 78.4 & 65.0 & 89.3 & 68.5 & 63.2 & 61.0 & 87.2 \\
        OPT-13B & 76.5 & 63.0 & 89.1 & 65.4 & 64.1 & 61.2 & 86.5 \\
        OPT-6.7B & 73.9 & 60.6 & 86.2 & 65.1 & 63.6 & 60.0 & 85.9 \\
        OPT-2.7B & 72.1 & 62.8 & 84.9 & 67.1 & 63.4 & 62.1 & 83.2 \\
        OPT-1.3B & 71.3 & 63.3 & 81.6 & 62.7 & 61.6 & 62.8 & 81.2 \\
        \bottomrule
    \end{tabular}
    \caption{The performance of the models on XSum with \benchmark alternative-choices using avg. PMI as the scoring function.}
    \label{tab:xsum_nfmg_avg_pmi}
\end{table*}

\begin{table*}[!ht]
    \small
    \centering
    \begin{tabular}{@{} l c c c c c c c @{}}
        \toprule
        Model & BART- & BART- & BLOOM- & distil- & distil- & PEGASUS & T5- \\
        & base & large & 560m & BART & PEGASUS &  & large\\
       \midrule
        T0-3B & 19.7 & 1.2 & 87.4 & 1.2 & 2.1 & 3.0 & 76.2 \\
        T0 & 33.9 & 5.3 & 80.3 & 5.4 & 5.7 & 3.2 & 84.1 \\
        FLAN-T5-xl & 19.2 & 2.4 & 85.7 & 4.9 & 3.4 & 3.4 & 74.5 \\
        FLAN-T5-xxl & 26.3 & 5.3 & 86.8 & 5.6 & 5.5 & 3.7 & 78.5 \\
        T5-LM-Adapt-xl & 19.7 & 9.7 & 40.9 & 12.4 & 11.7 & 15.8 & 51.1 \\
        T5-LM-Adapt-xxl & 23.8 & 8.9 & 51.2 & 12.0 & 10.1 & 9.6 & 61.3 \\
        GPT-Neo-1.3B & 26.3 & 10.9 & 31.4 & 21.2 & 14.2 & 13.7 & 50.5 \\
        GPT2-XL & 28.3 & 9.7 & 39.6 & 16.1 & 13.3 & 11.2 & 57.8 \\
        GPT-Neo-2.7B & 32.0 & 10.6 & 36.5 & 20.5 & 12.8 & 12.1 & 58.0 \\
        GPTJ-6B & 35.2 & 7.0 & 43.2 & 18.5 & 9.8 & 10.5 & 66.7 \\
        GPT-Neox-20B & 39.1 & 8.5 & 46.3 & 20.0 & 9.6 & 10.5 & 71.4 \\
        BLOOM & 42.8 & 8.5 & 50.9 & 20.7 & 9.8 & 10.7 & 72.5 \\
        BLOOM-7B1 & 32.6 & 10.9 & 43.0 & 20.7 & 13.3 & 13.9 & 60.9 \\
        BLOOM-3B & 30.5 & 13.8 & 39.8 & 19.8 & 18.3 & 18.7 & 51.3 \\
        BLOOM-1B7 & 27.0 & 14.7 & 36.9 & 22.9 & 19.2 & 21.5 & 44.1 \\
        BLOOM-1B1 & 24.8 & 17.1 & 35.2 & 24.9 & 21.7 & 24.7 & 40.6 \\
        OPT-175B & 48.8 & 8.7 & 56.0 & 20.7 & 9.8 & 7.8 & 78.9 \\
        OPT-66B & 44.3 & 8.2 & 50.7 & 19.8 & 9.2 & 7.3 & 77.6 \\
        OPT-30B & 45.6 & 7.7 & 50.7 & 20.7 & 9.6 & 8.4 & 76.6 \\
        OPT-13B & 41.0 & 8.7 & 47.8 & 18.8 & 9.4 & 8.7 & 73.7 \\
        OPT-6.7B & 37.1 & 8.0 & 43.4 & 17.8 & 8.2 & 8.7 & 69.6 \\
        OPT-2.7B & 33.7 & 8.7 & 39.6 & 21.0 & 10.3 & 10.5 & 67.7 \\
        OPT-1.3B & 29.8 & 8.5 & 37.7 & 17.6 & 11.2 & 10.7 & 62.3 \\
        \bottomrule
    \end{tabular}
    \caption{The performance of the models on XSum with \benchmark alternative-choices using avg. LL as the scoring function.}
    \label{tab:xsum_nfmg_avg_ll}
\end{table*}

\begin{table*}[!ht]
    \small
    \centering
    \begin{tabular}{@{} l c c c c c c c @{}}
        \toprule
        Model & BART- & BART- & BLOOM- & distil- & distil- & PEGASUS & T5- \\
        & base & large & 560m & BART & PEGASUS &  & large\\
       \midrule
        T0-3B & 48.8 & 26.1 & 83.2 & 27.3 & 29.7 & 27.4 & 91.1 \\
        T0 & 53.8 & 16.4 & 91.2 & 19.3 & 18.1 & 16.0 & 91.9 \\
        FLAN-T5-xl & 46.2 & 25.8 & 82.6 & 30.2 & 31.1 & 29.0 & 88.6 \\
        FLAN-T5-xxl & 54.6 & 30.9 & 85.7 & 34.4 & 36.6 & 33.6 & 89.9 \\
        T5-LM-Adapt-xl & 59.2 & 45.2 & 42.6 & 48.3 & 52.6 & 48.9 & 82.8 \\
        T5-LM-Adapt-xxl & 60.5 & 42.5 & 54.7 & 48.3 & 48.7 & 43.6 & 84.5 \\
        GPT-Neo-1.3B & 64.8 & 56.8 & 21.0 & 65.9 & 59.5 & 58.4 & 75.4 \\
        GPT2-XL & 61.8 & 49.0 & 33.3 & 57.1 & 53.8 & 54.1 & 74.9 \\
        GPT-Neo-2.7B & 63.9 & 51.7 & 23.9 & 60.2 & 55.1 & 55.7 & 76.2 \\
        GPTJ-6B & 70.0 & 49.0 & 28.9 & 66.6 & 54.7 & 54.1 & 80.7 \\
        GPT-Neox-20B & 68.5 & 51.0 & 29.4 & 65.6 & 55.8 & 53.4 & 82.6 \\
        BLOOM & 65.2 & 51.0 & 45.1 & 58.5 & 55.8 & 54.3 & 83.0 \\
        BLOOM-7B1 & 64.8 & 53.4 & 30.6 & 61.2 & 56.8 & 56.6 & 79.1 \\
        BLOOM-3B & 67.6 & 56.0 & 34.0 & 66.1 & 58.1 & 60.0 & 78.1 \\
        BLOOM-1B7 & 62.9 & 53.6 & 25.2 & 62.9 & 59.3 & 59.1 & 74.5 \\
        BLOOM-1B1 & 59.2 & 50.2 & 29.4 & 61.7 & 55.8 & 57.3 & 71.2 \\
        OPT-175B & 71.9 & 50.0 & 39.8 & 61.5 & 55.8 & 53.7 & 85.7 \\
        OPT-66B & 68.0 & 53.6 & 28.5 & 58.8 & 54.0 & 54.3 & 84.3 \\
        OPT-30B & 69.5 & 48.3 & 33.8 & 59.8 & 53.3 & 54.1 & 83.2 \\
        OPT-13B & 66.7 & 48.8 & 31.2 & 58.0 & 54.5 & 53.4 & 82.2 \\
        OPT-6.7B & 64.8 & 47.8 & 26.2 & 59.8 & 51.0 & 55.9 & 82.4 \\
        OPT-2.7B & 63.5 & 50.7 & 24.5 & 59.3 & 53.1 & 55.3 & 81.0 \\
        OPT-1.3B & 63.5 & 50.0 & 22.6 & 57.1 & 51.9 & 55.7 & 77.0 \\
        \bottomrule
    \end{tabular}
    \caption{The performance of the models on XSum with \benchmark alternative-choices using PMI as the scoring function.}
    \label{tab:xsum_nfmg_pmi}
\end{table*}

\begin{table*}[!ht]
    \small
    \centering
    \begin{tabular}{@{} l c c c c c c c @{}}
        \toprule
        Model & BART- & BART- & BLOOM- & distil- & distil- & PEGASUS & T5- \\
        & base & large & 560m & BART & PEGASUS &  & large\\
       \midrule
        T0-3B & 28.5 & 4.8 & 98.5 & 4.9 & 6.2 & 5.9 & 78.3 \\
        T0 & 42.8 & 10.4 & 98.7 & 8.3 & 7.3 & 5.9 & 84.9 \\
        FLAN-T5-xl & 30.5 & 8.9 & 98.7 & 6.8 & 7.8 & 8.7 & 74.5 \\
        FLAN-T5-xxl & 40.0 & 12.1 & 99.2 & 10.2 & 11.2 & 9.1 & 79.1 \\
        T5-LM-Adapt-xl & 39.1 & 29.7 & 97.3 & 26.3 & 26.1 & 27.6 & 58.2 \\
        T5-LM-Adapt-xxl & 42.1 & 24.2 & 97.7 & 23.2 & 20.1 & 21.2 & 65.8 \\
        GPT-Neo-1.3B & 44.3 & 31.2 & 96.2 & 36.3 & 28.6 & 27.6 & 56.7 \\
        GPT2-XL & 45.1 & 28.0 & 96.2 & 31.5 & 24.7 & 24.0 & 61.7 \\
        GPT-Neo-2.7B & 48.2 & 28.3 & 96.0 & 33.9 & 25.4 & 26.5 & 61.3 \\
        GPTJ-6B & 52.9 & 25.8 & 97.9 & 33.2 & 21.1 & 21.7 & 68.1 \\
        GPT-Neox-20B & 54.6 & 24.6 & 97.9 & 33.9 & 20.4 & 20.1 & 72.7 \\
        BLOOM & 54.0 & 26.1 & 98.1 & 32.4 & 23.6 & 22.1 & 73.7 \\
        BLOOM-7B1 & 49.2 & 30.2 & 97.5 & 33.4 & 28.8 & 29.7 & 62.1 \\
        BLOOM-3B & 44.3 & 33.8 & 96.4 & 34.6 & 31.6 & 34.7 & 57.8 \\
        BLOOM-1B7 & 45.1 & 34.8 & 96.0 & 37.8 & 32.7 & 34.7 & 52.2 \\
        BLOOM-1B1 & 44.1 & 37.7 & 94.8 & 39.5 & 34.6 & 37.4 & 51.3 \\
        OPT-175B & 59.0 & 23.4 & 98.3 & 30.5 & 17.4 & 16.4 & 80.5 \\
        OPT-66B & 57.0 & 24.2 & 98.3 & 30.2 & 19.2 & 14.4 & 77.6 \\
        OPT-30B & 55.7 & 22.9 & 97.9 & 30.2 & 18.3 & 16.0 & 77.2 \\
        OPT-13B & 51.8 & 23.2 & 98.1 & 28.8 & 18.5 & 17.8 & 75.4 \\
        OPT-6.7B & 52.7 & 23.7 & 97.1 & 29.3 & 18.1 & 16.7 & 71.4 \\
        OPT-2.7B & 49.9 & 26.1 & 97.3 & 30.2 & 19.7 & 19.9 & 67.3 \\
        OPT-1.3B & 45.8 & 26.6 & 97.1 & 30.5 & 22.4 & 23.1 & 62.5 \\
        \bottomrule
    \end{tabular}
    \caption{The performance of the models on XSum with \benchmark alternative-choices using LL as the scoring function.}
    \label{tab:xsum_nfmg_ll}
\end{table*}

\begin{table*}[!ht]
    \centering
    \tiny
    \begin{tabular}{@{} l c c c c c c c c c c c c c c c @{}}
        \toprule
        Model & B & BL & HG & L & MS & MI & NS & OD & O & PB & PT & R & RE & T & TS \\
       \midrule
        T0-3B & 11.5 & 0.0 & 9.1 & 20.0 & 4.8 & 11.8 & 20.8 & 51.4 & 13.0 & 0.0 & 0.0 & 25.8 & 4.2 & 13.9 & 15.2 \\
        T0 & 7.7 & 0.0 & 4.5 & 0.0 & 4.8 & 8.8 & 12.5 & 37.5 & 9.3 & 0.0 & 0.0 & 9.7 & 0.0 & 8.3 & 8.7 \\
        FLAN-T5-xl & 11.5 & 0.0 & 9.1 & 0.0 & 4.8 & 8.8 & 25.0 & 37.5 & 13.0 & 0.0 & 3.7 & 25.8 & 8.3 & 13.9 & 17.4 \\
        FLAN-T5-xxl & 11.5 & 0.0 & 9.1 & 0.0 & 4.8 & 8.8 & 16.7 & 37.5 & 7.4 & 0.0 & 0.0 & 19.4 & 8.3 & 8.3 & 17.4 \\
        T5-LM-Adapt-xl & 7.7 & 0.0 & 4.5 & 0.0 & 4.8 & 8.8 & 20.8 & 37.5 & 9.3 & 0.0 & 0.0 & 25.8 & 0.0 & 5.6 & 8.7 \\
        T5-LM-Adapt-xxl & 11.5 & 0.0 & 9.1 & 0.0 & 4.8 & 14.7 & 20.8 & 30.6 & 7.4 & 0.0 & 0.0 & 9.7 & 8.3 & 8.3 & 10.9 \\
        GPT-Neo-1.3B & 0.0 & 4.3 & 4.5 & 0.0 & 4.8 & 2.9 & 29.2 & 25.0 & 3.7 & 0.0 & 0.0 & 16.1 & 4.2 & 2.8 & 4.3 \\
        GPT2-XL & 3.8 & 0.0 & 0.0 & 0.0 & 4.8 & 2.9 & 33.3 & 27.8 & 3.7 & 0.0 & 0.0 & 25.8 & 4.2 & 2.8 & 4.3 \\
        GPT-Neo-2.7B & 7.7 & 0.0 & 9.1 & 0.0 & 4.8 & 5.9 & 29.2 & 23.6 & 3.7 & 0.0 & 0.0 & 16.1 & 8.3 & 5.6 & 8.7 \\
        GPTJ-6B & 0.0 & 4.3 & 0.0 & 0.0 & 4.8 & 2.9 & 29.2 & 22.2 & 3.7 & 0.0 & 0.0 & 9.7 & 4.2 & 5.6 & 6.5 \\
        GPT-Neox-20B & 11.5 & 0.0 & 9.1 & 20.0 & 9.5 & 5.9 & 20.8 & 23.6 & 5.6 & 0.0 & 0.0 & 16.1 & 8.3 & 5.6 & 8.7 \\
        BLOOM & 11.5 & 4.3 & 9.1 & 0.0 & 9.5 & 5.9 & 16.7 & 19.4 & 5.6 & 8.3 & 0.0 & 12.9 & 8.3 & 5.6 & 4.3 \\
        BLOOM-7B1 & 7.7 & 8.7 & 0.0 & 0.0 & 4.8 & 2.9 & 20.8 & 25.0 & 9.3 & 0.0 & 0.0 & 12.9 & 8.3 & 5.6 & 10.9 \\
        BLOOM-3B & 3.8 & 4.3 & 0.0 & 0.0 & 4.8 & 2.9 & 16.7 & 20.8 & 5.6 & 0.0 & 0.0 & 9.7 & 4.2 & 5.6 & 8.7 \\
        BLOOM-1B7 & 3.8 & 4.3 & 4.5 & 0.0 & 4.8 & 2.9 & 20.8 & 25.0 & 7.4 & 0.0 & 0.0 & 12.9 & 8.3 & 2.8 & 6.5 \\
        BLOOM-1B1 & 3.8 & 4.3 & 9.1 & 20.0 & 4.8 & 5.9 & 25.0 & 23.6 & 7.4 & 8.3 & 3.7 & 16.1 & 12.5 & 5.6 & 8.7 \\
        OPT-175B & 7.7 & 4.3 & 9.1 & 40.0 & 9.5 & 8.8 & 12.5 & 23.6 & 5.6 & 8.3 & 0.0 & 9.7 & 8.3 & 8.3 & 10.9 \\
        OPT-66B & 7.7 & 4.3 & 9.1 & 0.0 & 9.5 & 5.9 & 12.5 & 20.8 & 7.4 & 0.0 & 0.0 & 6.5 & 8.3 & 8.3 & 8.7 \\
        OPT-30B & 7.7 & 4.3 & 9.1 & 0.0 & 4.8 & 5.9 & 16.7 & 19.4 & 5.6 & 0.0 & 0.0 & 9.7 & 8.3 & 8.3 & 8.7 \\
        OPT-13B & 7.7 & 0.0 & 9.1 & 0.0 & 9.5 & 5.9 & 16.7 & 26.4 & 3.7 & 0.0 & 0.0 & 12.9 & 8.3 & 5.6 & 6.5 \\
        OPT-6.7B & 7.7 & 4.3 & 4.5 & 20.0 & 4.8 & 5.9 & 16.7 & 23.6 & 5.6 & 8.3 & 0.0 & 6.5 & 12.5 & 5.6 & 10.9 \\
        OPT-2.7B & 7.7 & 4.3 & 4.5 & 0.0 & 4.8 & 8.8 & 16.7 & 25.0 & 3.7 & 0.0 & 0.0 & 12.9 & 8.3 & 5.6 & 8.7 \\
        OPT-1.3B & 3.8 & 4.3 & 4.5 & 0.0 & 4.8 & 5.9 & 20.8 & 19.4 & 5.6 & 0.0 & 0.0 & 12.9 & 8.3 & 2.8 & 4.3 \\
        \bottomrule
    \end{tabular}
    \caption{The performance of the models on CNN/DM with \benchmark alternative-choices using avg. PMI as the scoring function. The models are   BanditSumm (B), BERT\_LSTM\_PN\_RL (BL), Heter-Graph (HG), Lead3 (L), MatchSumm (MS), MI-unsup (MI), NeuSumm (NS), Oracle (discourse) (OD), Oracle (O), Pacsum (bert) (PB), Pacsum (tfidf) (PT), Refresh (R), RNN\_Ext\_RL (RE), Textrank (T), Textrank (st) (TS) \\}
    \label{tab:cnn_dm_nfmg_avg_pmi}
\end{table*}

\begin{table*}[!ht]
    \centering
    \tiny
    \begin{tabular}{@{} l c c c c c c c c c c c c c c c @{}}
        \toprule
        Model & B & BL & HG & L & MS & MI & NS & OD & O & PB & PT & R & RE & T & TS \\
       \midrule
        T0-3B & 0.0 & 0.0 & 0.0 & 0.0 & 0.0 & 5.9 & 12.5 & 33.3 & 7.4 & 0.0 & 3.7 & 25.8 & 0.0 & 8.3 & 17.4 \\
        T0 & 0.0 & 0.0 & 0.0 & 0.0 & 0.0 & 2.9 & 8.3 & 23.6 & 9.3 & 0.0 & 3.7 & 12.9 & 0.0 & 5.6 & 13.0 \\
        FLAN-T5-xl & 0.0 & 0.0 & 0.0 & 0.0 & 0.0 & 8.8 & 12.5 & 25.0 & 5.6 & 0.0 & 0.0 & 12.9 & 0.0 & 8.3 & 15.2 \\
        FLAN-T5-xxl & 0.0 & 0.0 & 0.0 & 0.0 & 0.0 & 2.9 & 4.2 & 18.1 & 3.7 & 0.0 & 0.0 & 6.5 & 0.0 & 5.6 & 15.2 \\
        T5-LM-Adapt-xl & 0.0 & 0.0 & 0.0 & 0.0 & 0.0 & 2.9 & 4.2 & 18.1 & 7.4 & 0.0 & 3.7 & 9.7 & 0.0 & 2.8 & 13.0 \\
        T5-LM-Adapt-xxl & 0.0 & 0.0 & 0.0 & 0.0 & 0.0 & 2.9 & 4.2 & 12.5 & 5.6 & 0.0 & 3.7 & 6.5 & 0.0 & 2.8 & 13.0 \\
        GPT-Neo-1.3B & 0.0 & 0.0 & 0.0 & 0.0 & 0.0 & 0.0 & 4.2 & 4.2 & 0.0 & 0.0 & 0.0 & 0.0 & 0.0 & 2.8 & 2.2 \\
        GPT2-XL & 0.0 & 0.0 & 0.0 & 0.0 & 0.0 & 0.0 & 4.2 & 5.6 & 3.7 & 0.0 & 0.0 & 6.5 & 0.0 & 2.8 & 4.3 \\
        GPT-Neo-2.7B & 0.0 & 0.0 & 0.0 & 0.0 & 0.0 & 0.0 & 4.2 & 5.6 & 0.0 & 0.0 & 0.0 & 3.2 & 0.0 & 2.8 & 2.2 \\
        GPTJ-6B & 0.0 & 0.0 & 0.0 & 0.0 & 0.0 & 0.0 & 4.2 & 5.6 & 1.9 & 0.0 & 0.0 & 0.0 & 0.0 & 2.8 & 4.3 \\
        GPT-Neox-20B & 0.0 & 0.0 & 0.0 & 0.0 & 0.0 & 0.0 & 0.0 & 5.6 & 1.9 & 0.0 & 0.0 & 0.0 & 0.0 & 2.8 & 4.3 \\
        BLOOM & 0.0 & 0.0 & 0.0 & 0.0 & 0.0 & 0.0 & 0.0 & 4.2 & 1.9 & 0.0 & 0.0 & 0.0 & 0.0 & 2.8 & 6.5 \\
        BLOOM-7B1 & 0.0 & 0.0 & 0.0 & 0.0 & 0.0 & 0.0 & 4.2 & 8.3 & 3.7 & 0.0 & 0.0 & 0.0 & 0.0 & 2.8 & 6.5 \\
        BLOOM-3B & 0.0 & 0.0 & 0.0 & 0.0 & 0.0 & 2.9 & 4.2 & 4.2 & 1.9 & 0.0 & 0.0 & 0.0 & 0.0 & 2.8 & 6.5 \\
        BLOOM-1B7 & 0.0 & 0.0 & 0.0 & 0.0 & 0.0 & 0.0 & 4.2 & 6.9 & 0.0 & 0.0 & 0.0 & 0.0 & 0.0 & 2.8 & 6.5 \\
        BLOOM-1B1 & 0.0 & 0.0 & 0.0 & 0.0 & 0.0 & 2.9 & 4.2 & 5.6 & 1.9 & 0.0 & 0.0 & 3.2 & 0.0 & 2.8 & 6.5 \\
        OPT-175B & 0.0 & 0.0 & 0.0 & 0.0 & 0.0 & 2.9 & 4.2 & 4.2 & 1.9 & 0.0 & 0.0 & 3.2 & 0.0 & 2.8 & 4.3 \\
        OPT-66B & 0.0 & 0.0 & 0.0 & 0.0 & 0.0 & 2.9 & 4.2 & 5.6 & 1.9 & 0.0 & 0.0 & 3.2 & 0.0 & 2.8 & 2.2 \\
        OPT-30B & 0.0 & 0.0 & 0.0 & 0.0 & 0.0 & 0.0 & 4.2 & 5.6 & 1.9 & 0.0 & 0.0 & 3.2 & 0.0 & 2.8 & 2.2 \\
        OPT-13B & 0.0 & 0.0 & 0.0 & 0.0 & 0.0 & 0.0 & 4.2 & 4.2 & 1.9 & 0.0 & 0.0 & 3.2 & 0.0 & 2.8 & 2.2 \\
        OPT-6.7B & 0.0 & 0.0 & 0.0 & 0.0 & 0.0 & 2.9 & 4.2 & 8.3 & 1.9 & 0.0 & 0.0 & 3.2 & 0.0 & 2.8 & 2.2 \\
        OPT-2.7B & 0.0 & 0.0 & 0.0 & 0.0 & 0.0 & 2.9 & 4.2 & 6.9 & 1.9 & 0.0 & 0.0 & 3.2 & 0.0 & 2.8 & 4.3 \\
        OPT-1.3B & 0.0 & 0.0 & 0.0 & 0.0 & 0.0 & 2.9 & 4.2 & 4.2 & 0.0 & 0.0 & 0.0 & 6.5 & 0.0 & 2.8 & 2.2 \\
        \bottomrule
    \end{tabular}
    \caption{The performance of the models on CNN/DM with \benchmark alternative-choices using avg. LL as the scoring function. The models are   BanditSumm (B), BERT\_LSTM\_PN\_RL (BL), Heter-Graph (HG), Lead3 (L), MatchSumm (MS), MI-unsup (MI), NeuSumm (NS), Oracle (discourse) (OD), Oracle (O), Pacsum (bert) (PB), Pacsum (tfidf) (PT), Refresh (R), RNN\_Ext\_RL (RE), Textrank (T), Textrank (st) (TS) \\}
    \label{tab:cnn_dm_nfmg_avg_ll}
\end{table*}

\begin{table*}[!ht]
    \centering
    \tiny
    \begin{tabular}{@{} l c c c c c c c c c c c c c c c @{}}
        \toprule
        Model & B & BL & HG & L & MS & MI & NS & OD & O & PB & PT & R & RE & T & TS \\
       \midrule
        T0-3B & 0.0 & 0.0 & 0.0 & 0.0 & 4.8 & 0.0 & 8.3 & 33.3 & 13.0 & 0.0 & 0.0 & 9.7 & 0.0 & 2.8 & 2.2 \\
        T0 & 0.0 & 0.0 & 0.0 & 0.0 & 4.8 & 0.0 & 0.0 & 22.2 & 13.0 & 0.0 & 0.0 & 3.2 & 0.0 & 2.8 & 4.3 \\
        FLAN-T5-xl & 0.0 & 0.0 & 0.0 & 0.0 & 4.8 & 0.0 & 8.3 & 25.0 & 13.0 & 0.0 & 0.0 & 12.9 & 0.0 & 0.0 & 2.2 \\
        FLAN-T5-xxl & 0.0 & 0.0 & 0.0 & 0.0 & 4.8 & 0.0 & 8.3 & 20.8 & 11.1 & 0.0 & 0.0 & 3.2 & 0.0 & 0.0 & 4.3 \\
        T5-LM-Adapt-xl & 0.0 & 0.0 & 0.0 & 0.0 & 4.8 & 0.0 & 4.2 & 25.0 & 11.1 & 0.0 & 0.0 & 3.2 & 0.0 & 0.0 & 2.2 \\
        T5-LM-Adapt-xxl & 0.0 & 0.0 & 0.0 & 0.0 & 4.8 & 0.0 & 0.0 & 22.2 & 9.3 & 0.0 & 0.0 & 3.2 & 0.0 & 0.0 & 2.2 \\
        GPT-Neo-1.3B & 0.0 & 0.0 & 0.0 & 0.0 & 4.8 & 0.0 & 4.2 & 16.7 & 5.6 & 0.0 & 0.0 & 0.0 & 0.0 & 0.0 & 0.0 \\
        GPT2-XL & 0.0 & 0.0 & 0.0 & 0.0 & 4.8 & 0.0 & 0.0 & 13.9 & 5.6 & 0.0 & 0.0 & 6.5 & 0.0 & 0.0 & 0.0 \\
        GPT-Neo-2.7B & 0.0 & 0.0 & 0.0 & 0.0 & 4.8 & 0.0 & 0.0 & 11.1 & 5.6 & 0.0 & 0.0 & 0.0 & 0.0 & 0.0 & 0.0 \\
        GPTJ-6B & 0.0 & 0.0 & 0.0 & 0.0 & 0.0 & 0.0 & 4.2 & 11.1 & 3.7 & 0.0 & 0.0 & 0.0 & 0.0 & 0.0 & 0.0 \\
        GPT-Neox-20B & 0.0 & 0.0 & 0.0 & 0.0 & 4.8 & 0.0 & 0.0 & 8.3 & 5.6 & 0.0 & 0.0 & 3.2 & 0.0 & 0.0 & 0.0 \\
        BLOOM & 0.0 & 0.0 & 0.0 & 0.0 & 4.8 & 0.0 & 0.0 & 9.7 & 3.7 & 0.0 & 0.0 & 3.2 & 0.0 & 0.0 & 0.0 \\
        BLOOM-7B1 & 0.0 & 0.0 & 0.0 & 0.0 & 4.8 & 0.0 & 4.2 & 11.1 & 5.6 & 0.0 & 0.0 & 0.0 & 0.0 & 0.0 & 0.0 \\
        BLOOM-3B & 0.0 & 0.0 & 0.0 & 0.0 & 4.8 & 0.0 & 0.0 & 12.5 & 5.6 & 0.0 & 0.0 & 0.0 & 0.0 & 0.0 & 0.0 \\
        BLOOM-1B7 & 0.0 & 0.0 & 0.0 & 0.0 & 4.8 & 0.0 & 0.0 & 16.7 & 3.7 & 0.0 & 0.0 & 0.0 & 0.0 & 0.0 & 0.0 \\
        BLOOM-1B1 & 0.0 & 0.0 & 0.0 & 0.0 & 4.8 & 0.0 & 4.2 & 15.3 & 5.6 & 0.0 & 0.0 & 0.0 & 0.0 & 0.0 & 0.0 \\
        OPT-175B & 0.0 & 0.0 & 0.0 & 0.0 & 4.8 & 0.0 & 0.0 & 11.1 & 5.6 & 0.0 & 0.0 & 3.2 & 0.0 & 0.0 & 0.0 \\
        OPT-66B & 0.0 & 0.0 & 0.0 & 0.0 & 4.8 & 0.0 & 0.0 & 9.7 & 5.6 & 0.0 & 0.0 & 0.0 & 0.0 & 0.0 & 0.0 \\
        OPT-30B & 0.0 & 0.0 & 0.0 & 0.0 & 4.8 & 0.0 & 0.0 & 12.5 & 7.4 & 0.0 & 0.0 & 0.0 & 0.0 & 0.0 & 0.0 \\
        OPT-13B & 0.0 & 0.0 & 0.0 & 0.0 & 4.8 & 0.0 & 0.0 & 13.9 & 5.6 & 0.0 & 0.0 & 0.0 & 0.0 & 0.0 & 0.0 \\
        OPT-6.7B & 0.0 & 0.0 & 0.0 & 0.0 & 4.8 & 0.0 & 0.0 & 9.7 & 5.6 & 0.0 & 0.0 & 0.0 & 0.0 & 0.0 & 0.0 \\
        OPT-2.7B & 0.0 & 0.0 & 0.0 & 0.0 & 4.8 & 0.0 & 0.0 & 13.9 & 5.6 & 0.0 & 0.0 & 0.0 & 0.0 & 0.0 & 0.0 \\
        OPT-1.3B & 0.0 & 0.0 & 0.0 & 0.0 & 4.8 & 0.0 & 0.0 & 16.7 & 7.4 & 0.0 & 0.0 & 0.0 & 0.0 & 0.0 & 0.0 \\
        \bottomrule
    \end{tabular}
    \caption{The performance of the models on CNN/DM with \benchmark alternative-choices using PMI as the scoring function. The models are   BanditSumm (B), BERT\_LSTM\_PN\_RL (BL), Heter-Graph (HG), Lead3 (L), MatchSumm (MS), MI-unsup (MI), NeuSumm (NS), Oracle (discourse) (OD), Oracle (O), Pacsum (bert) (PB), Pacsum (tfidf) (PT), Refresh (R), RNN\_Ext\_RL (RE), Textrank (T), Textrank (st) (TS) \\}
    \label{tab:cnn_dm_nfmg_pmi}
\end{table*}

\begin{table*}[!ht]
    \centering
    \tiny
    \begin{tabular}{@{} l c c c c c c c c c c c c c c c @{}}
        \toprule
        Model & B & BL & HG & L & MS & MI & NS & OD & O & PB & PT & R & RE & T & TS \\
       \midrule
        T0-3B & 26.9 & 21.7 & 45.5 & 40.0 & 42.9 & 61.8 & 75.0 & 81.9 & 29.6 & 33.3 & 48.1 & 74.2 & 54.2 & 44.4 & 54.3 \\
        T0 & 15.4 & 21.7 & 31.8 & 0.0 & 38.1 & 58.8 & 62.5 & 65.3 & 16.7 & 8.3 & 40.7 & 61.3 & 37.5 & 47.2 & 50.0 \\
        FLAN-T5-xl & 23.1 & 30.4 & 31.8 & 20.0 & 47.6 & 61.8 & 75.0 & 68.1 & 18.5 & 16.7 & 40.7 & 74.2 & 58.3 & 47.2 & 56.5 \\
        FLAN-T5-xxl & 7.7 & 17.4 & 18.2 & 0.0 & 23.8 & 47.1 & 50.0 & 68.1 & 13.0 & 0.0 & 18.5 & 45.2 & 29.2 & 41.7 & 47.8 \\
        T5-LM-Adapt-xl & 38.5 & 47.8 & 36.4 & 0.0 & 38.1 & 64.7 & 70.8 & 65.3 & 14.8 & 16.7 & 29.6 & 61.3 & 37.5 & 41.7 & 47.8 \\
        T5-LM-Adapt-xxl & 34.6 & 26.1 & 9.1 & 0.0 & 23.8 & 55.9 & 54.2 & 52.8 & 13.0 & 0.0 & 11.1 & 48.4 & 20.8 & 27.8 & 37.0 \\
        GPT-Neo-1.3B & 11.5 & 13.0 & 9.1 & 0.0 & 14.3 & 41.2 & 62.5 & 19.4 & 3.7 & 0.0 & 7.4 & 45.2 & 12.5 & 16.7 & 23.9 \\
        GPT2-XL & 23.1 & 17.4 & 9.1 & 0.0 & 19.0 & 47.1 & 66.7 & 25.0 & 9.3 & 0.0 & 18.5 & 54.8 & 29.2 & 22.2 & 28.3 \\
        GPT-Neo-2.7B & 15.4 & 17.4 & 4.5 & 0.0 & 14.3 & 44.1 & 50.0 & 19.4 & 5.6 & 0.0 & 7.4 & 38.7 & 16.7 & 16.7 & 23.9 \\
        GPTJ-6B & 7.7 & 21.7 & 4.5 & 0.0 & 14.3 & 41.2 & 41.7 & 25.0 & 3.7 & 0.0 & 7.4 & 35.5 & 4.2 & 13.9 & 23.9 \\
        GPT-Neox-20B & 0.0 & 13.0 & 4.5 & 0.0 & 14.3 & 47.1 & 50.0 & 26.4 & 3.7 & 0.0 & 14.8 & 32.3 & 4.2 & 25.0 & 28.3 \\
        BLOOM & 7.7 & 13.0 & 4.5 & 0.0 & 14.3 & 38.2 & 29.2 & 20.8 & 5.6 & 0.0 & 3.7 & 29.0 & 16.7 & 13.9 & 21.7 \\
        BLOOM-7B1 & 19.2 & 17.4 & 0.0 & 0.0 & 9.5 & 44.1 & 45.8 & 22.2 & 7.4 & 0.0 & 11.1 & 41.9 & 16.7 & 19.4 & 26.1 \\
        BLOOM-3B & 23.1 & 13.0 & 4.5 & 0.0 & 19.0 & 44.1 & 50.0 & 22.2 & 3.7 & 0.0 & 3.7 & 41.9 & 16.7 & 16.7 & 23.9 \\
        BLOOM-1B7 & 19.2 & 17.4 & 9.1 & 0.0 & 9.5 & 44.1 & 41.7 & 26.4 & 3.7 & 0.0 & 3.7 & 41.9 & 12.5 & 16.7 & 21.7 \\
        BLOOM-1B1 & 23.1 & 26.1 & 4.5 & 0.0 & 19.0 & 44.1 & 54.2 & 22.2 & 5.6 & 0.0 & 11.1 & 41.9 & 25.0 & 25.0 & 23.9 \\
        OPT-175B & 23.1 & 34.8 & 4.5 & 0.0 & 19.0 & 50.0 & 45.8 & 19.4 & 3.7 & 0.0 & 7.4 & 32.3 & 16.7 & 16.7 & 21.7 \\
        OPT-66B & 30.8 & 30.4 & 4.5 & 0.0 & 19.0 & 47.1 & 54.2 & 22.2 & 7.4 & 0.0 & 11.1 & 38.7 & 12.5 & 19.4 & 30.4 \\
        OPT-30B & 26.9 & 34.8 & 4.5 & 0.0 & 14.3 & 50.0 & 45.8 & 20.8 & 3.7 & 0.0 & 3.7 & 35.5 & 12.5 & 13.9 & 26.1 \\
        OPT-13B & 30.8 & 30.4 & 4.5 & 0.0 & 19.0 & 47.1 & 54.2 & 29.2 & 5.6 & 0.0 & 11.1 & 38.7 & 25.0 & 16.7 & 23.9 \\
        OPT-6.7B & 30.8 & 39.1 & 4.5 & 0.0 & 19.0 & 50.0 & 58.3 & 26.4 & 7.4 & 0.0 & 18.5 & 38.7 & 29.2 & 27.8 & 26.1 \\
        OPT-2.7B & 23.1 & 26.1 & 4.5 & 0.0 & 28.6 & 50.0 & 58.3 & 33.3 & 7.4 & 0.0 & 11.1 & 45.2 & 16.7 & 19.4 & 26.1 \\
        OPT-1.3B & 26.9 & 30.4 & 9.1 & 0.0 & 19.0 & 44.1 & 62.5 & 25.0 & 7.4 & 0.0 & 7.4 & 45.2 & 16.7 & 16.7 & 23.9 \\
        \bottomrule
    \end{tabular}
    \caption{The performance of the models on CNN/DM with \benchmark alternative-choices using LL as the scoring function. The models are   BanditSumm (B), BERT\_LSTM\_PN\_RL (BL), Heter-Graph (HG), Lead3 (L), MatchSumm (MS), MI-unsup (MI), NeuSumm (NS), Oracle (discourse) (OD), Oracle (O), Pacsum (bert) (PB), Pacsum (tfidf) (PT), Refresh (R), RNN\_Ext\_RL (RE), Textrank (T), Textrank (st) (TS) \\}
    \label{tab:cnn_dm_nfmg_ll}
\end{table*}

\section{Accuracies from Models Used to Generate Summaries}
\label{sec:model_trained}
We show the performance of different models using the same models to generate the alternative summaries for XSum using different scoring functions in \cref{tab:xsum_trained}.

\begin{table*}[!ht]
    \small
    \centering
    \begin{tabular}{@{} l c c c c c c c c @{}}
        \toprule
        Model & Scoring & BART- & BART- & BLOOM- & distil- & distil- & PEGASUS & T5- \\
        & Function & base & large & 560m & BART & PEGASUS &  & large\\
       \midrule
        BART-base & Avg. PMI & 24.4 & 42.5 & 95.4 & 34.4 & 45.1 & 42.2 & 83.0 \\
        BART-base & Avg. LL & 0.0 & 2.2 & 97.1 & 0.5 & 3.4 & 5.5 & 50.1 \\
        BART-base & PMI & 17.7 & 26.6 & 64.8 & 27.1 & 35.0 & 34.7 & 77.4 \\
        BART-base & LL & 0.6 & 8.9 & 99.6 & 2.0 & 8.9 & 13.5 & 54.5 \\
        BART-large & Avg. PMI & 63.5 & 24.4 & 96.0 & 29.5 & 39.4 & 32.2 & 94.2 \\
        BART-large & Avg. LL & 32.8 & 0.0 & 96.9 & 4.4 & 2.5 & 3.0 & 77.0 \\
        BART-large & PMI & 52.9 & 17.9 & 62.3 & 26.8 & 32.3 & 29.2 & 91.1 \\
        BART-large & LL & 42.8 & 1.0 & 99.6 & 7.3 & 4.8 & 5.7 & 77.6 \\
        BLOOM-560m & Avg. PMI & 55.9 & 44.7 & 52.8 & 53.9 & 45.8 & 46.1 & 72.0 \\
        BLOOM-560m & Avg. LL & 18.6 & 6.0 & 0.4 & 11.7 & 6.6 & 7.5 & 50.9 \\
        BLOOM-560m & PMI & 49.5 & 36.5 & 10.7 & 48.3 & 40.7 & 42.2 & 68.9 \\
        BLOOM-560m & LL & 32.2 & 16.7 & 37.3 & 21.5 & 12.8 & 14.8 & 57.8 \\
        distil-BART & Avg. PMI & 51.0 & 24.2 & 94.5 & 16.6 & 35.7 & 30.8 & 93.4 \\
        distil-BART & Avg. LL & 11.0 & 0.0 & 97.7 & 0.0 & 2.1 & 4.3 & 72.5 \\
        distil-BART & PMI & 44.7 & 18.6 & 52.8 & 18.8 & 30.9 & 26.5 & 88.6 \\
        distil-BART & LL & 20.7 & 1.7 & 99.6 & 0.0 & 4.6 & 7.3 & 73.1 \\
        distil-PEGASUS & Avg. PMI & 62.9 & 34.1 & 97.3 & 32.4 & 19.7 & 18.9 & 94.8 \\
        distil-PEGASUS & Avg. LL & 16.4 & 1.9 & 88.9 & 2.0 & 0.0 & 0.7 & 74.1 \\
        distil-PEGASUS & PMI & 51.4 & 22.7 & 77.8 & 26.6 & 17.2 & 17.1 & 92.3 \\
        distil-PEGASUS & LL & 27.0 & 5.6 & 98.5 & 3.9 & 0.2 & 1.8 & 76.2 \\
        PEGASUS & Avg. PMI & 72.4 & 44.9 & 97.1 & 42.9 & 36.4 & 22.8 & 96.9 \\
        PEGASUS & Avg. LL & 29.4 & 1.7 & 87.8 & 2.9 & 0.5 & 0.0 & 84.3 \\
        PEGASUS & PMI & 65.4 & 29.7 & 79.9 & 37.3 & 26.8 & 19.2 & 94.2 \\
        PEGASUS & LL & 38.9 & 5.8 & 99.0 & 7.8 & 2.3 & 0.2 & 85.3 \\
        T5-large & Avg. PMI & 43.2 & 50.7 & 93.5 & 46.1 & 51.5 & 49.8 & 31.7 \\
        T5-large & Avg. LL & 8.6 & 12.3 & 94.8 & 10.2 & 13.3 & 18.9 & 0.2 \\
        T5-large & PMI & 34.1 & 34.5 & 59.3 & 36.3 & 42.1 & 42.0 & 27.7 \\
        T5-large & LL & 28.5 & 31.9 & 99.2 & 26.1 & 28.4 & 34.2 & 4.1 \\
        \bottomrule
    \end{tabular}
    \caption{The performance of the models on XSum using the same models to generate the \fic{} summary.}
    \label{tab:xsum_trained}
\end{table*}

\end{document}